\documentclass{article}

\PassOptionsToPackage{numbers, compress}{natbib}

\usepackage[main,final]{neurips_2026}

\usepackage[utf8]{inputenc} 
\usepackage[T1]{fontenc}    
\usepackage{hyperref}       
\usepackage{url}            
\usepackage{booktabs}       
\usepackage{amsfonts}       
\usepackage{nicefrac}       
\usepackage{microtype}      
\usepackage{xcolor}         
\usepackage{graphicx}       
\usepackage{subcaption}     
\usepackage{float}
\usepackage{amsmath,amssymb}
\usepackage{amsthm}

\usepackage{enumitem}

\theoremstyle{plain}
\newtheorem{theorem}{Theorem}
\newtheorem{proposition}{Proposition}
\newtheorem{corollary}{Corollary}

\theoremstyle{definition}

\providecommand{\demaybefig}[2][\linewidth]{%
  \IfFileExists{#2}{%
    \includegraphics[width=#1]{#2}%
  }{%
    \fbox{\parbox[c][1cm][c]{0.92\linewidth}{\centering\ttfamily\small missing: \detokenize{#2}}}%
  }%
}

\title{Macrocanonical Generator Networks:\\ Data-Efficient Neural Surrogates for Amortized Physics Simulation}

\author{%
  Niall Jeffrey \\
  Dept.\ of Physics \& \\ King's Institute for Artificial Intelligence \\ King's College London \\ \texttt{niall.jeffrey@kcl.ac.uk}
  \And
  Benjamin D. Wandelt \\
  Dept.\ of Physics and Astronomy \& \\ Dept.\ of Applied Mathematics and Statistics \\ Johns Hopkins University \\ \texttt{wandelt@jhu.edu}
}

\begin{document}

\maketitle
\begin{abstract}
Many scientific simulations are computationally expensive, limiting their use in simulation-based inference, uncertainty quantification, and decision making. We introduce \emph{macrocanonical generator networks}, a framework for learning fast, data-efficient neural surrogates for \emph{amortized} simulation of stationary physical processes by matching multiscale statistics of a target process. Inspired by microcanonical maximum-entropy synthesis, which matches feature statistics per sample by inner-loop optimization, our approach instead trains a neural generator to satisfy the same constraints in expectation, preserving physically realistic sample-to-sample fluctuations. We prove that gradient-descent training induces preconditioned gradient-descent dynamics in sample space that inherit the symmetry-preservation properties of the microcanonical gradient descent of Bruna and Mallat, amortizing its per-sample inner-loop optimization into a single training run after which every sample is generated in one forward pass. We further show that residual generator parameterizations initialized near the identity, trained with a small learning rate, preserve entropy locally during early optimization and mitigate mode collapse. Experiments on cosmology and fluid simulations show that macrocanonical generators recover realistic samples from extremely small training sets, in some cases a single training example, while generating new realizations more than $10^6$ times faster per sample than the reference simulator and more than $10^3$ times faster than microcanonical gradient descent. The trained generators can be reused for downstream pretraining, including simulation-based inference, providing a data-efficient route to multi-fidelity scientific machine learning workflows.
\end{abstract}

\section{Introduction}
\vspace{-0.3cm}

\begin{figure}[!t]
  \centering
  \demaybefig{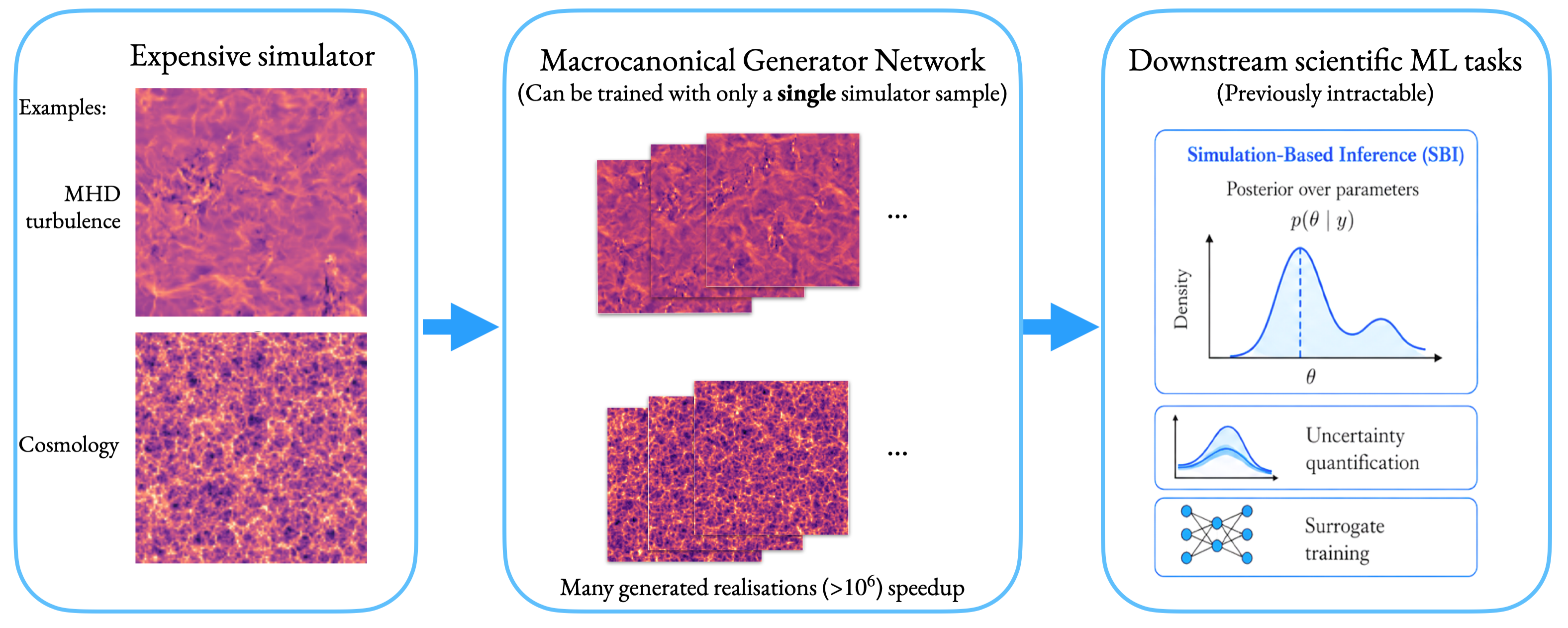}
  \caption{Macrocanonical generator networks (\emph{centre}) turn a small set of expensive simulator outputs (\emph{left}; turbulence and cosmology targets, Section~\ref{sec:experiments}) into large ensembles of new realizations matching the chosen statistical features (e.g. wavelet phase harmonics) of the targets, unlocking downstream scientific machine-learning tasks (\emph{right}) that are intractable with the expensive simulator alone.}
  \label{fig:overview}
\vspace{-0.6cm}
\end{figure}

Scientific simulations are essential tools for understanding complex physical systems.
However, high-fidelity simulators are often computationally expensive, sometimes requiring days to generate a single realization because each sample typically requires an iterative solve or stochastic rollout.
This limits their use in tasks such as simulation-based inference \cite{alsing2019fast,cranmer2020frontier}, uncertainty quantification, and decision making, where large numbers of simulations are required.
Examples include cosmological simulation \cite{villaescusa2023camels,jeffrey2025dark}, tsunami models \cite{behrens2015tsunami}, wind farm simulation \cite{kirby2023wind}, and nuclear fusion models \cite{hoppe2021dream}.

Recent advances in deep generative modelling have demonstrated that neural networks can learn complex data distributions. Methods such as generative adversarial networks \cite{goodfellow2014generative}, diffusion models \cite{ho2020denoising}, and normalizing flows \cite{rezende2015variational} have achieved remarkable success in image generation and related domains. Earlier work such as Generative Moment Matching Networks \cite{li2015gmmn} trains generators by matching the empirical data distribution using the maximum mean discrepancy (MMD). However, these methods typically rely on large training datasets and are not designed for settings in which only a small number of simulator realizations are available, a regime that is common in scientific computing.

Other moment-matching approaches, such as MMD-GAN \cite{li2017mmdgan}, the method of simulated moments \cite{mcfadden1989msm} and turbulence closures \cite{pope2000turbulent}, as well as neural operators (which address the different task of predicting a simulator's output for given boundary conditions), likewise require many simulations or observations, whereas we match a fixed multiscale moment vector (features) that a single stationary realization can be sufficient to estimate.

Many physical processes of interest---turbulent flows, cosmological fields, multifractal processes---are \emph{stationary}: their statistical properties are invariant under spatial translations.
This stationarity is a powerful inductive bias.
As shown by Bruna and Mallat \cite{bruna2018multiscale}, the translation symmetry of a stationary process means that a single realization of dimension $d$ provides $O(d)$ translated copies of local structures, enabling accurate estimation of statistical moments from as few as a single example.

In such systems, knowledge about the process is naturally expressed through \emph{statistical constraints}.
Multiscale statistics derived from wavelet transforms, scattering transforms \cite{allys2020new,cheng2024scattering}, or wavelet phase harmonics \cite{mallat2020phase,zhang2021maximum} can capture important non-Gaussian structure in complex signals.
Throughout, ``multiscale'' refers to spatial scales rather than temporal, and all signals considered are equal-time snapshots.
Classical texture synthesis methods exploit this structure by generating samples that match these statistics \cite{portilla2000parametric}.
In these approaches, samples are synthesized individually by optimizing pixel values until the target statistics are satisfied.
This procedure can be interpreted as a form of microcanonical maximum-entropy sampling \cite{bruna2018multiscale,zhang2021maximum}, where the probability distribution is conditioned on the empirical statistics of the observed signal.
Related recent work extends moment-constrained generation with diffusion-based samplers that guide moments along stochastic trajectories toward maximum-entropy models \cite{lempereur2026mgd}.
These approaches remain per-sample iterative at generation time.

In this work we introduce \emph{macrocanonical generator networks}, a neural framework that generalizes this idea.
Instead of synthesizing each realization independently through optimization, we train a neural generator that produces samples whose statistics match those of the target process \emph{in expectation}.
The generator therefore learns a global transport map that produces samples satisfying the desired statistical constraints.
Moreover, enforcing constraints in expectation rather than per sample is a better match to real physical processes, in which individual realizations fluctuate around their defining statistics (Section~\ref{sec:macro}).

This shift from per-sample optimization to learning a global generator has important consequences.
Once trained, the generator acts as a fast surrogate simulator capable of producing large numbers of realizations in one forward pass per sample, effectively amortizing the expense of iterative simulation and iterative moment-guided sampling.
In this sense, our objective is complementary to trajectory-based approaches: rather than improving a per-sample iterative sampler, we learn a reusable transport map for high-throughput workflows.
Because the training objective depends on statistical constraints rather than individual examples, and because stationarity provides abundant constraints from few realizations, the method can operate effectively in extremely data-limited regimes.
The resulting generators can be used to pretrain neural networks for downstream tasks, enabling new approaches to simulation-based inference \cite{alsing2019fast,cranmer2020frontier}, multi-fidelity modelling \cite{hikida2025multilevel}, and the training of scientific foundation models \cite{parker2025aion,mccabe2025walrus,wiesner2025physics}.

Beyond the practical advantages, the macrocanonical formulation provides a conceptual bridge between classical statistical synthesis methods and modern deep generative modelling.
In particular, we show that training such generators induces a preconditioned gradient descent dynamics in sample space, closely related to the optimization procedure used in microcanonical synthesis \cite{bruna2018multiscale,zhang2021maximum}.
Two design choices are key: (i)~residual generator parameterizations initialized near the identity, and (ii)~sufficiently low learning rates.
Together, these help the generator preserve entropy locally during early optimization and mitigate collapse to degenerate outputs.
Without these inductive biases, the generator can collapse toward a constant mapping $F(x) \approx y^\ast$, effectively giving the same output for every input, which is a situation of near-minimal entropy.

\begin{table}[H]     
  \centering                                           
  \small       
  \caption{Macrocanonical generator networks (ours) compared with prevailing deep generative methods and with classical microcanonical synthesis. Properties marked ``exact, by construction'' follow from architectural equivariance and the loss; properties marked ``approximate (learned)'' hold only insofar as the training set is large and diverse enough to make them statistically apparent.}                                                                 
  \label{tab:method-comparison}      
   \resizebox{\textwidth}{!}{%
   \begin{tabular}{lcccc}                                    
     \toprule                              
      & \textbf{Macrocanonical} & \textbf{GAN}  & \textbf{Diffusion} & \textbf{Microcanonical} \\                  
      & \textbf{(ours)}         &              & \textbf{model}     & \textbf{synthesis}     \\                    
     \midrule                                                                                                      
    Training samples needed         & $\ge 1$ realization        & large dataset       & large dataset       & $\ge 1$ realization      \\
    Inference cost per sample       & 1 forward pass             & 1 forward pass      & many ($\sim 10^{2\text{--}3}$) steps & inner-loop optimization \\
    Symmetries of moment constraints & exact, by construction    & approximate (learned) & approximate (learned) & exact, by construction  \\                
    Per-realization fluctuations    & yes (constraints in expectation) & yes           & yes                & no (Dirac on $\phi^\ast$) \\                   
    Training signal                 & feature-moment matching    & adversarial         & denoising score matching & per-sample feature matching \\           
    Distributional flexibility      & limited to constrained moments & unrestricted    & unrestricted        & limited to constrained moments \\             
    \bottomrule                                                                                                   
  \end{tabular}%
  }
\end{table} 
The main contributions of this work are therefore:

(i)~We introduce macrocanonical generator networks, a generative framework that learns neural transport maps producing samples whose feature statistics match those of a target process while replacing per-sample iterative simulation or sampling with amortized one-pass generation after training.

(ii)~We establish a connection between neural generator training and microcanonical synthesis, showing that generator training induces preconditioned gradient descent dynamics.

(iii)~We analyze the entropy behavior of residual generators and show that identity-initialized residual parameterizations with low learning rates help to preserve local volume and maintain diversity in generated samples, while generators without these inductive biases suffer from mode collapse.

(iv)~We demonstrate that macrocanonical generators can learn effective surrogate simulators in extremely data-limited regimes, including cases with a single training example, leveraging the stationarity of the target process as a key inductive bias and making many-query analyses with surrogate simulators computationally practical.

(v)~We show how such generators can be used for pretraining in simulation-based inference pipelines, enabling new multi-fidelity workflows and providing a potential foundation for training scientific foundation models.

Together, these results suggest that macrocanonical generator networks provide a new framework for constructing fast neural surrogate simulators of stationary physical processes, bridging ideas from statistical physics, classical synthesis methods, and modern deep generative modelling.

Table~\ref{tab:method-comparison} compares macrocanonical generators with the most widely used deep generative model classes and classical microcanonical synthesis.                                                      
The framework trades distributional generality for three properties that are important in the data-limited scientific regime: (i) training from a handful, or even a single, realization; (ii) by-construction preservation of every 
symmetry respected by the moment constraints, rather than approximate symmetry learned from data; and (iii) amortized one-pass generation, like a GAN and unlike a diffusion model.

\section{Microcanonical and Macrocanonical Generation}
\vspace{-0.3cm}

We first formalize the distinction between microcanonical synthesis methods and the macrocanonical generator formulation introduced in this work.

\subsection{Microcanonical synthesis}
\vspace{-0.3cm}

Let $\phi(x)$ denote a differentiable vector of feature statistics extracted from a signal $x$.
Given a reference signal with statistics $\phi^\ast$, microcanonical synthesis generates samples by minimizing
\begin{equation}
L(x) = \frac12 \|\phi(x) - \phi^\ast\|^2 .
\end{equation}
Starting from an initialization $x^{(0)}$ drawn from a stationary Gaussian distribution with periodic boundary conditions \cite{bruna2018multiscale,zhang2021maximum}, samples are obtained by gradient descent in signal space
\begin{equation}
x^{(t+1)} = x^{(t)} - \eta \nabla_x L(x^{(t)}).
\end{equation}
Each synthesized sample is therefore produced by an independent optimization procedure that progressively adjusts the signal until the desired feature constraints are satisfied.
When the feature statistics $\phi$ are shift-equivariant (as is the case for wavelet-based statistics), the gradient descent preserves the stationarity of the initial distribution \cite{bruna2018multiscale}.
The Gaussian initialization provides maximum entropy subject to a mean and variance constraint, and convergence to the microcanonical set is guaranteed when the step size lies below the inverse Lipschitz constant of the gradient \cite{bruna2018multiscale}.
The resulting distribution is not in general maximum entropy on the microcanonical set, but inherits symmetries of the initial measure including stationarity (Theorem~3.4 and Corollary~3.5 of \citealt{bruna2018multiscale}).

\subsection{Macrocanonical generators}
\vspace{-0.3cm}
\label{sec:macro}

The terminology reflects a physical distinction.
A microcanonical formulation requires each realization to satisfy $\phi = \phi^\ast$ exactly, eliminating fluctuations. This is generally unphysical, because real physical systems are generated by ensembles defined only through expectation values.
In the macrocanonical formulation the ensemble fixes $\phi^\ast = \langle\phi\rangle$, so individual realizations fluctuate around the constraint, as observed for spectra and analogous features in quantum field configurations, cosmological fields, turbulence, and quantum many-body energy spectra.

Instead of optimizing each sample independently, we introduce a neural generator $y = F_\theta(x)$, where $x \sim p_0$ is drawn from a simple input distribution such as a stationary Gaussian field with periodic boundary conditions.
The generator is trained so that feature constraints are satisfied in expectation:
\begin{equation}
\mathbb{E}_{x \sim p_0}[\phi(F_\theta(x))] \approx \phi^\ast .
\end{equation}
This corresponds to a macrocanonical formulation \cite{jaynes1957information,cover2006elements}: moment constraints are enforced on the ensemble of generated samples rather than on each individual sample.
The trained network therefore learns a global transport map $F_\theta : x \mapsto y$ that produces samples satisfying the same statistical constraints that would otherwise be enforced by per-sample optimization.
Crucially, once training is complete, generating a new sample requires only a single forward pass through the network, whereas microcanonical synthesis requires a full iterative optimization for every sample.
The cost of learning the transport map is thus amortized over all subsequent samples.

\paragraph{Why few realizations still work:}
\label{sec:few-realizations}
The claim that $\phi^\ast$ can be estimated from very few realizations, or even a single realization, is a consequence of stationarity together with the shift-invariant, spatially averaged form of the feature statistics \cite{bruna2018multiscale}.
Each wavelet phase harmonic coefficient is an average over translates, $\phi_k(x)=\tfrac{1}{d}\sum_{u}f_k(T_u x)$, of a local functional $f_k$, where $T_u$ denotes translation by $u$ on a periodic grid of $d$ sites.
For a stationary field the variance of this average is $\tfrac{1}{d}\sum_{u}\mathrm{Cov}\bigl(f_k(X),f_k(T_uX)\bigr)$, which is $O(1/d)$ if the spatial correlations of $f_k$ decay sufficiently rapidly. A single realization then behaves like an average over $O(d)$ weakly correlated copies, and $n$ independent realizations reduce the variance to $O(1/(nd))$, so field size and the number of realizations play the same statistical role \cite{bruna2018multiscale}.

\subsection{Residual generator parameterizations}
\vspace{-0.3cm}
\label{sec:residual}

We consider generator architectures that are initialized near the identity map \cite{he2016deep}.
In particular we study residual parameterizations of the form
\begin{equation}
\label{eq:additive}
F_\theta(x) = x + G_\theta(x)
\end{equation}
and
\begin{equation}
\label{eq:multiplicative}
F_\theta(x) = G_\theta(x) + H_\theta(x) \odot x,
\end{equation}
where $G_\theta$ and $H_\theta$ are convolutional networks.
The choice of convolutional architecture is motivated by the analogy with microcanonical synthesis: in microcanonical gradient descent, shift-equivariant features $\phi$ ensure that the gradient transport preserves stationarity \cite{bruna2018multiscale,zhang2021maximum}.
Similarly, convolutional $G_\theta$ and $H_\theta$ make the generator $F_\theta$ translation-equivariant, relating the macrocanonical generator to this symmetry-preserving structure.
These parameterizations are initialized so that
\begin{equation}
G_{\theta_0}(x) = 0, \quad H_{\theta_0}(x) = \mathbf{1}.
\end{equation}
Under this initialization the generator initially acts as the identity mapping: $F_{\theta_0}(x) = x$.
If the input $x$ is drawn from a stationary distribution, the generated samples are therefore also stationary at initialization.
During early stages of optimization with a low learning rate, the learned transport remains close to the identity, which plays an important role in preserving both the entropy and the stationarity of the generated distribution, as we analyze in the following sections.

\section{Training Dynamics of Macrocanonical Generators}
\vspace{-0.3cm}
\label{sec:dynamics}

We now analyze how gradient-based training of the generator affects the generated samples.
Let $y=F_\theta(x)$ for $x\sim p_0$, and define the macrocanonical training objective
\begin{equation}
L(\theta)
=
\frac12
\left\|
\mathbb E_{x\sim p_0}[\phi(F_\theta(x))]-\phi^\ast
\right\|^2.
\end{equation}
To understand the effect of this update in sample space, we linearize the generator around the current parameters.

\begin{theorem}[Generator training induces preconditioned gradient descent]
\label{thm:preconditioned}
Assume (i) the generator $F_\theta$ is twice continuously differentiable with respect to its parameters $\theta$, (ii) the feature map $\phi$ is differentiable with respect to its input, (iii) the learning rate $\eta$ is sufficiently small that the second-order remainder below is negligible relative to the first-order term, and (iv) the required expectations and derivatives exist and differentiation may be interchanged with expectation.

Let $L(\theta)$ denote the macrocanonical objective defined above{. Write $m(\theta)=\mathbb{E}_{x''\sim p_0}[\phi(F_\theta(x''))]$ for the mean feature vector, so $L(\theta)=\tfrac12\|m(\theta)-\phi^\ast\|^2$.
For each input $x'$, define a per-sample objective
\begin{equation}
\widetilde L_{x'}(y)
\;=\;
\bigl\langle \phi(y),\; m(\theta^{(t)})-\phi^\ast \bigr\rangle .
\end{equation}
}
Then, for a fixed input sample $x$, the corresponding change in the generated sample satisfies, to first order in the parameter update,
\begin{equation}
y^{(t+1)}(x)
\approx
y^{(t)}(x)
-
\eta
\mathbb{E}_{x' \sim p_0}
\left[
K_{\theta^{(t)}}(x,x')\, \nabla_y \widetilde L_{x'}
\right],
\end{equation}
where $y^{(t)}(x)=F_{\theta^{(t)}}(x)$, {the approximation error is $O(\|\Delta\theta\|^{2})=O(\eta^{2})$ by assumption (i)},
\begin{equation}
K_{\theta^{(t)}}(x,x')
=
J_{\theta^{(t)}}(x) J_{\theta^{(t)}}(x')^\top,
\qquad
J_{\theta^{(t)}}(x)
=
\frac{\partial F_{\theta^{(t)}}(x)}{\partial \theta}.
\end{equation} {$\nabla_y \widetilde L_{x'}$ denotes the gradient of the per-sample objective evaluated at the current output $y = F_{\theta^{(t)}}(x')$},
\begin{equation}
\nabla_y \widetilde L_{x'}
=
\left(\nabla_y \phi(F_{\theta^{(t)}}(x'))\right)^\top
\left(
{m(\theta^{(t)})} - \phi^\ast
\right).
\end{equation}
Thus generator training induces a preconditioned gradient descent in sample space, governed by the kernel $K_{\theta^{(t)}}(x,x')$.
\end{theorem}

\paragraph{Kernel interpretation.}
The full sample-space update is coupled across inputs through the cross-kernel $K_\theta(x,x') = J_\theta(x)J_\theta(x')^\top$.
The update can therefore be written
\begin{equation}
y^{(t+1)} = y^{(t)} - \eta K_\theta \nabla_y \widetilde L ,
\end{equation}
where $K_\theta$ acts as an integral operator on the gradient field
\begin{equation}
(K_\theta g)(x)
=
\mathbb{E}_{x' \sim p_0}\!\left[K_\theta(x,x') g(x')\right].
\end{equation}
Generator training therefore induces a kernel-preconditioned gradient descent in sample space.
The kernel couples updates across inputs through the generator Jacobian, so that the evolution of each sample depends on the sample-space gradients evaluated at other inputs $x'$.
{We note that for a batch containing the single input $x$, the update of the output at that same input reduces to a pointwise preconditioned step with the self-kernel $K_\theta(x,x)=J_\theta(x)J_\theta(x)^\top$, analogous to the Gauss--Newton or neural tangent kernel preconditioner \cite{jacot2018neural}.}

The theorem does not assume training on a single (noise) input $x$: the update $\Delta\theta=-\eta\nabla_\theta L(\theta^{(t)})$ uses the full macrocanonical objective, with the expectation over $x'$ inside. The theorem gives the change in the output at arbitrary input $x$.
Since all outputs share $\theta$, one update changes every sample.

\begin{proof}[Proof sketch]
The argument has two ingredients.
First, by {assumptions~(i) and~(iii)} the parameter update $\Delta\theta = -\eta\,\nabla_\theta L(\theta^{(t)})$ is small, so we may linearize the generator around $\theta^{(t)}$,
\begin{equation}
F_{\theta^{(t)}+\Delta\theta}(x) \approx F_{\theta^{(t)}}(x) + J_{\theta^{(t)}}(x)\,\Delta\theta,
\qquad J_{\theta^{(t)}}(x) = \frac{\partial F_{\theta^{(t)}}(x)}{\partial \theta},
\end{equation}
which converts the parameter update into a sample-space update $y^{(t+1)}(x) \approx y^{(t)}(x) - \eta\, J_{\theta^{(t)}}(x)\,\nabla_\theta L(\theta^{(t)})$.
Second, the chain rule applied to $L(\theta) = \tfrac12 \|m(\theta) - \phi^\ast\|^2$ with $m(\theta) = \mathbb{E}_{x'\sim p_0}[\phi(F_\theta(x'))]$, together with the regularity assumptions (ii) and (iv), expresses
\begin{equation}
\nabla_\theta L(\theta^{(t)}) = \mathbb{E}_{x' \sim p_0}\bigl[ J_{\theta^{(t)}}(x')^\top\, \nabla_y \widetilde L_{x'} \bigr],
\end{equation}
with $\nabla_y \widetilde L_{x'}$ as defined in the theorem statement.
Substituting this expectation into the linearized sample update and identifying $K_{\theta^{(t)}}(x,x') = J_{\theta^{(t)}}(x)\,J_{\theta^{(t)}}(x')^\top$ gives the stated kernel-preconditioned form.
A complete derivation, with the chain-rule manipulations written out step by step, is given in Appendix~\ref{app:proof-thm1}.
\end{proof}

\begin{proposition}[Symmetry preservation under equivariant generators]
\label{prop:symmetry}
Let $\mathcal{G}$ be a group acting on the sample space by measurable transformations.
Let $X \sim p_0$ and define $Y = F_\theta(X)$.
Assume that: \\
(1)  $p_0$ is $\mathcal{G}$-invariant, i.e.\ for all $g \in \mathcal{G}$, the random variable $gX$ has the same distribution as $X$; \\
(2) the generator is $\mathcal{G}$-equivariant, i.e.\ $F_\theta(gx)=gF_\theta(x)$ for all $g \in \mathcal{G}$.

Then $Y$ is $\mathcal{G}$-invariant: for all $g \in \mathcal{G}$, the random variables $gY$ and $Y$ have the same distribution.
\end{proposition}

\begin{proof}
Fix $g \in \mathcal{G}$.
Since $p_0$ is $\mathcal{G}$-invariant:
$gX \stackrel{d}{=} X$.

Since $gX \stackrel{d}{=} X$ and $F_\theta$ is measurable,
\begin{equation}
F_\theta(gX) \stackrel{d}{=} F_\theta(X)=Y.
\end{equation}
By equivariance of the generator,
\begin{equation}
F_\theta(gX)=gF_\theta(X)=gY, \mathrm{\ therefore\ }
gY \stackrel{d}{=} Y.
\end{equation}
Since this holds for every $g \in \mathcal{G}$, the random variable $Y$ is $\mathcal{G}$-invariant.
\end{proof}

\begin{corollary}[Stationarity of macrocanonical samples]
\label{cor:stationarity}
Let $\mathcal{G}$ be the group of periodic translations on the grid.
If $p_0$ is stationary and $F_\theta$ is translation-equivariant (as enforced by a convolutional architecture with periodic boundary conditions), then $p_\theta=\mathcal{L}(F_\theta(X))$ is stationary.
\end{corollary}

\begin{proof}
Take $\mathcal{G}$ to be the group of periodic translations on the grid.
By assumption, $p_0$ is invariant under these translations, and $F_\theta$ is equivariant under the same action.
Proposition~\ref{prop:symmetry} therefore implies that $F_\theta(X)$ is stationary.
\end{proof}

\begin{corollary}[Symmetry inheritance and amortization of microcanonical synthesis]
\label{cor:micro-inheritance}
{Assume the conditions of Theorem~\ref{thm:preconditioned}, let $\mathcal{G}$ be the group of periodic grid translations acting by permutation matrices $g$, and assume $\phi$ is shift-invariant ($\phi(gy)=\phi(y)$ for all $g\in\mathcal{G}$), $F_\theta$ is translation-equivariant ($F_\theta(gx)=gF_\theta(x)$; e.g., a convolutional network with periodic boundary conditions), and $p_0$ is stationary.
Then, for every $g \in \mathcal{G}$ and all inputs $x, x'$:
(i)~the per-sample gradient is equivariant, $\nabla_y \widetilde L_{gx'} = g\, \nabla_y \widetilde L_{x'}$;
(ii)~the kernel intertwines the translation action, $K_{\theta^{(t)}}(gx, gx') = g\, K_{\theta^{(t)}}(x,x')\, g^{-1}$;
and (iii)~consequently, the one-step update map $U^{(t)}: y^{(t)}(\cdot) \mapsto y^{(t+1)}(\cdot)$ of Theorem~\ref{thm:preconditioned} commutes with translations, $U^{(t)}(g\,\cdot) = g\, U^{(t)}(\cdot)$, and the generated distribution of $y^{(t)}$ is stationary at every training step $t$.

Stationarity of the generated distribution in fact holds exactly, for every parameter value, by Corollary~\ref{cor:stationarity}; the content of (i)--(iii) is that the first-order training dynamics of Theorem~\ref{thm:preconditioned} respect the same symmetry.
Property (iii) is the same symmetry-preservation guarantee that \citet{bruna2018multiscale} establish for microcanonical pixel-space gradient descent (their Theorem~3.4), transferred to the training dynamics of the macrocanonical generator.
After a single training run the generator amortizes the per-sample inner-loop optimization required by microcanonical synthesis~\cite{bruna2018multiscale,zhang2021maximum}: every subsequent sample is produced in one forward pass while inheriting the stationarity-preservation guarantee of property (iii).}
\end{corollary}
\noindent\emph{Proof.} See Appendix~\ref{app:proof-cor2}.

Together, Theorem~\ref{thm:preconditioned} and Corollary~\ref{cor:micro-inheritance} provide the macrocanonical analogue of symmetry-preserving microcanonical dynamics.
Theorem~\ref{thm:preconditioned} describes the local sample-space geometry induced by training under the macrocanonical objective, in the regime of low learning rate where the generator remains well approximated by its local linearization.
Proposition~\ref{prop:symmetry} and Corollary~\ref{cor:stationarity} show that equivariant generators preserve symmetries of the input distribution, including stationarity.

When $K_\theta(x,x)$ is close to the identity, the local induced dynamics resemble the pixel-space gradient descent used in microcanonical synthesis \cite{bruna2018multiscale,zhang2021maximum}.

This connection reveals the importance of the learning rate $\eta$: the linearization underlying Theorem~\ref{thm:preconditioned} is valid when $\eta$ is sufficiently small.
This mirrors the requirement in microcanonical gradient descent that the step size be sufficiently small to ensure stable convergence.
In the macrocanonical setting, a low learning rate typically ensures that each training step produces a small perturbation of the transport map, helping preserve entropy and maintain the intended stationarity (in practice, numerically) of the generated distribution.
\begin{figure}[t]
\centering
\demaybefig{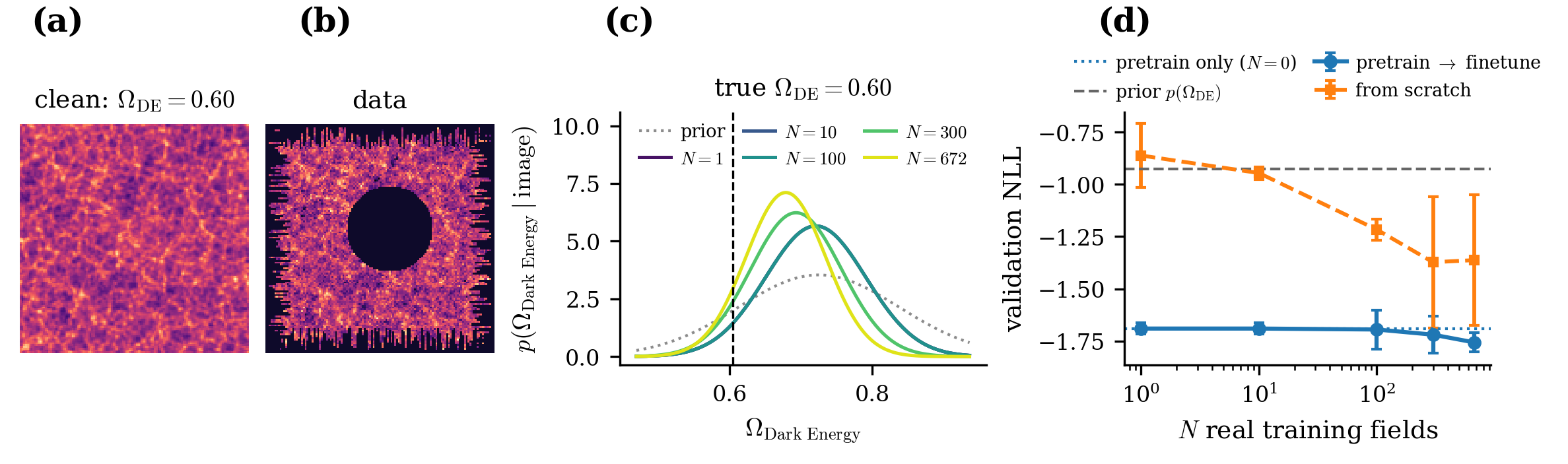}
\caption{Simulation-based inference of the dark-energy fraction $\Omega_{\rm DE}$ from 2-D density images (Section~\ref{sec:sbi}). \textbf{(a, b)} A validation example before and after the observational effects (i.e. masking). \textbf{(c)} Posterior after fine-tuning on $N$ ``real'' simulations. \textbf{(d)} Validation NLL against $N$ (inc. standard deviation with 3 seeds) using cosmologies in the $\Omega_{\rm DE}$ training range. The surrogate-pretrained model improves on the prior before any ``real'' simulation is used and consistently outperforms the baseline.}
\label{fig:sbi-summary}
\end{figure}

\section{Entropy and Volume Distortion}
\vspace{-0.3cm}
\label{sec:entropy}

We now examine how the generator transformation affects the diversity and stationarity of generated samples.
Let $x \sim p_0$ denote the input distribution and $y = F_\theta(x)$ the generated samples.

For a differentiable mapping $F_\theta$, the input Jacobian $\mathcal{J}_\theta(x) = \partial F_\theta(x)/\partial x$ describes how local volumes are distorted by the transformation.
We use $\mathcal{J}$ for the input Jacobian to distinguish it from the parameter Jacobian $J_\theta(x) = \partial F_\theta(x)/\partial \theta$ of Section~\ref{sec:dynamics}.
When $F_\theta$ is a diffeomorphism, the change of variables formula relates the differential entropy of the output to that of the input:
\begin{equation}
\mathcal{H}(Y) = \mathcal{H}(X) + \mathbb{E}_{x \sim p_0}\bigl[\log\bigl|\det \mathcal{J}_\theta(x)\bigr|\bigr],
\end{equation}
where $\mathcal{H}(\cdot)$ denotes differential entropy.
Even when $F_\theta$ is not globally invertible, the log absolute determinant of the Jacobian remains a useful measure of local volume change: regions where $|\det \mathcal{J}_\theta(x)| \approx 1$ are approximately volume-preserving, while $|\det \mathcal{J}_\theta(x)| \to 0$ indicates that the mapping is collapsing a neighborhood to a lower-dimensional set.

For residual generators $F_\theta(x) = x + G_\theta(x)$, the input Jacobian takes the form
\begin{equation}
\mathcal{J}_\theta(x) = I + \frac{\partial G_\theta(x)}{\partial x}.
\end{equation}
Because the generator is initialized with $G_{\theta_0}(x) = 0$, at initialization $\mathcal{J}_{\theta_0}(x) = I$ and therefore $\det \mathcal{J}_{\theta_0}(x) = 1$.
The entropy of the generated distribution equals that of the input: $\mathcal{H}(Y) = \mathcal{H}(X)$.
During early training with a low learning rate, the residual branch $G_\theta(x)$ can remain small, so the Jacobian tends to stay close to the identity and the mapping approximately preserves volume in input space.

This behavior helps maintain high entropy in the generated distribution and reduces the risk of mode collapse.
In contrast, a generator without these inductive biases can rapidly learn a mapping that contracts input space, driving $|\det \mathcal{J}_\theta(x)| \to 0$ and producing a degenerate distribution.
{In the extreme case, training can drive the generator toward a near-constant mapping $F_\theta(x) \approx y^\ast$ for all $x$, in which the generator outputs essentially the same image for every input and the generated distribution has near-minimal entropy.}

\begin{figure}[t]
  \centering
  \demaybefig[0.7\linewidth]{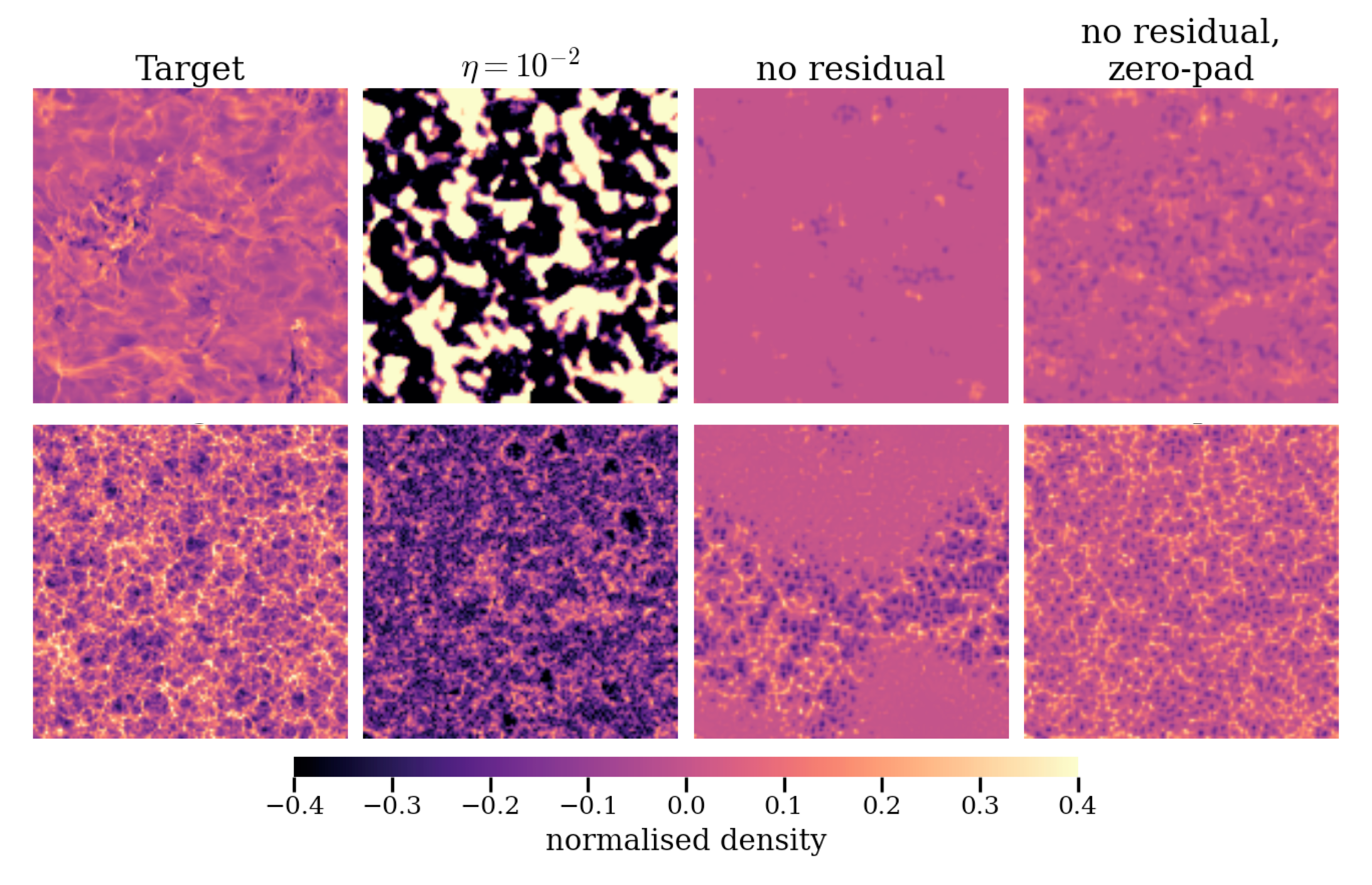}
  \caption{Inductive-bias ablation on the MHD (top) and cosmic-web (bottom) benchmarks; one target slice and one synthesis from each variant after $100$ optimizer steps. The three failure modes correspond to removing the small step ($\eta=10^{-2}$), the residual shortcut, and (additionally) the circular-padded convolutions; \emph{zero-pad} means the U-Net is no longer translation-equivariant, so Corollary~\ref{cor:stationarity} no longer guarantees stationarity (producing boundary-correlated structure).}
  \label{fig:ablation-inductive-bias}
\end{figure}

{
If the fixed output satisfies the constraints, $\phi(y^\ast)=\phi^\ast$ (i.e., $y^\ast$ is a microcanonical solution), the constant mapping $x \mapsto y^\ast$ attains zero training loss and is therefore a global minimizer.
Avoiding collapse is therefore a property of the training dynamics (identity initialization, residual parameterization and a small learning rate) rather than of the loss; empirically, the trained generator matches the across-sample variance of held-out simulations (Figure~\ref{fig:wph-fluctuations}), whereas removing these inductive biases produces collapse (Figure~\ref{fig:ablation-inductive-bias}).}

A further consequence concerns stationarity.
As noted in Section~\ref{sec:residual}, the convolutional structure of $G_\theta$ and $H_\theta$ makes the generator translation-equivariant, mirroring the role of shift-equivariant features in microcanonical gradient descent where stationarity is preserved \cite{bruna2018multiscale,zhang2021maximum}.
Together with the residual parameterization and low learning rate, which keep the transport close to the identity, these design choices help preserve the stationarity of the generated field throughout training.
Without these inductive biases, the generator can exhibit severe collapse, reducing entropy and sample diversity even when constrained moments are matched; in practical architectures this may yield apparent or effective stationarity violations in finite samples, even though exact equivariance would preserve stationarity in theory.
This failure mode is particularly damaging because stationarity is the symmetry that enables estimation of statistics from a single realization in data-limited settings.
Viewed through this lens, identity initialization and low learning rates act as an implicit regularization through the training dynamics, discouraging rapid collapse while feature constraints are being enforced.

To summarize, residual parameterizations initialized near the identity therefore provide two interrelated structural biases: (i)~local volume preservation, maintaining entropy and avoiding mode collapse; and (ii)~gradual departure from the identity, {keeping training in the small-update regime of Theorem~\ref{thm:preconditioned}, in which the induced dynamics retain the per-sample transport structure of microcanonical synthesis}.
Combined with the convolutional architecture, which encodes translation equivariance, these biases help maintain both the diversity and the stationarity of the generated distribution.
In practice, this avoids rapid mode collapse.

{We find that both of these structural biases are needed. Removing either the residual shortcut or the small learning rate produces collapse (Figure~\ref{fig:ablation-inductive-bias}), with detailed sweeps in Appendix~\ref{app:sweeps}.}

\section{Experiments}
\vspace{-0.3cm}
\label{sec:experiments}

We test macrocanonical generators on the axes that matter for scientific workflows: extreme data limits, faithful per-realization variability of summary statistics, wall-clock feasibility relative to reference simulators and iterative synthesis, and cold-start amortized inference when simulation budgets are inadequate.
The two periodic stationary targets are a single $128{\times}128$ magnetohydrodynamic (MHD) turbulence snapshot ($N_\text{train}=1$) and up to $100$ fiducial Quijote cosmic-web slices (Appendix~\ref{app:quijote}); Figure~\ref{fig:overview} shows targets and syntheses.
{MHD turbulence produces stationary, strongly non-Gaussian fields with structure coupled across scales. The cosmic web is the filamentary matter distribution formed by gravitational collapse, whose physical structures depend on the cosmological parameters.
In both cases, the simulators are often too expensive to generate the required training set sizes.}

\paragraph{Setup.}
\label{sec:setup}
We instantiate~\eqref{eq:additive}--\eqref{eq:multiplicative} with a four-stage circular-padded U-Net, so $F_\theta$ is translation-equivariant on the torus (Corollary~\ref{cor:stationarity}) and begins at the identity.
Features $\phi$ are wavelet phase harmonics~\cite{mallat2020phase,zhang2021maximum}; $\phi^\ast$ is the empirical mean of $\phi$ over the available training fields; inputs $x\sim p_0$ are stationary Gaussian fields with power spectrum matched to the training pool, so second-order structure is correct before higher-order moments shape $F_\theta$.
We optimize with Adam at $\eta=10^{-4}$ and global gradient-norm clipping.
Architecture, WPH settings and optimizer schedules are detailed in Appendices~\ref{app:experiment-details} and~\ref{app:wph}.

{\paragraph{Effect of training-set size.}
Trained on $1$, $10$ or $100$ fields from a separate suite of MHD turbulence simulations, scored against held-out runs (Appendix~\ref{app:mhd64}), the generator produces new fields rather than copies even from a single training field, and its statistics approach those of held-out simulations within about ten fields, as expected from Section~\ref{sec:few-realizations}.
On the cosmic-web benchmark, a single training field already reproduces the held-out sample-to-sample scatter.

\paragraph{Comparison with generative baselines.}
On the same data, WGAN-GP \cite{gulrajani2017improved}, flow matching \cite{lipman2023flow} and a DDPM \cite{ho2020denoising} memorize the training field when only one is available, even with translation augmentation \cite{favero2025compositional}, and close the gap only as many more fields are added (Appendix~\ref{app:mhd64}).
Microcanonical synthesis \cite{bruna2018multiscale,zhang2021maximum} removes sample-to-sample fluctuations and is significantly slower per sample.}

\paragraph{Per-realization feature distributions.}
\label{sec:cosmo}
Matching $\phi^\ast$ in expectation does not, by itself, guarantee that each realization exhibits realistic scatter in every summary channel.
Yet simulation-based inference (SBI) fits conditional densities by exploiting precisely those fluctuations: posteriors become miscalibrated if every surrogate has near-identical statistics.
Training on $\phi^\ast$ averaged over $100$ Quijote slices without supervising per-realization variance, we draw $100$ synthetic fields and compare each WPH coefficient against $100$ held-out simulator slices.
Coefficient-wise Kolmogorov--Smirnov tests give mean $p$-value $\langle p\rangle=0.52$, consistent with draws from the same marginals (Appendix~\ref{app:feature-dist}, Figure~\ref{fig:wph-fluctuations}), in line with Section~\ref{sec:entropy}.
Microcanonical synthesis~\cite{bruna2018multiscale,zhang2021maximum} instead forces $\phi=\phi^\ast$ per realization, collapsing to a Dirac mass and removing real likelihood contributions from the posterior density.

\paragraph{Surrogate simulation speed.}
\label{sec:surrogate-speed}
Cosmological inference ensembles routinely spend $\mathcal{O}(\text{hours})$--$\mathcal{O}(\text{days})$ of cluster time per nonlinear $N$-body realization~\cite{jeffrey2025dark}, and Quijote-class simulations inherit the same bottleneck.
After a single offline fit, our generator emits batches of fresh $128{\times}128$ fields at $\mathcal{O}(\text{ms})$ per sample on one NVIDIA A100---more than six orders of magnitude higher throughput than reference simulators at comparable fidelity, and more than $4{\times}10^3$ times faster than iterative microcanonical gradient-descent synthesis~\cite{bruna2018multiscale,zhang2021maximum} (which additionally destroys per-realization fluctuations, as noted above).
Once trained, the surrogate amortizes across arbitrarily many downstream queries---ensemble forecasting, uncertainty loops, iterative refinement---workflows otherwise inaccessible.
Per-iteration training cost and exact wall-clock numbers are tabulated in Appendix~\ref{app:throughput-sbi}.

\paragraph{Pretraining for simulation-based inference.}
\label{sec:sbi}
We infer $\Omega_{\rm DE}=1-\Omega_m$ from $128{\times}128$ matter-density projections in the Quijote Latin-hypercube suite.
{$\Omega_{\rm DE}$ is the dark-energy fraction of the cosmic energy-density, which causes the accelerated expansion of the Universe, the effect of which is imprinted on the  matter density structures (e.g. the morphology of filaments and halos). Posteriors on $\Omega_{\rm DE}$ from such projected density images are a simplified version of the inference problem faced by galaxy surveys.}
Each cosmology requires a full $N$-body simulation (about $800$ CPU core-hours per Quijote realization~\cite{villaescusa2020quijote}), so conventional neural SBI hits an obstacle: even thirty cosmologies {provide insufficient paired supervision for an image encoder} without auxiliary simulators.
We fit one macrocanonical generator per training cosmology from a single clean projection and synthesize a much larger labelled pool against which a small ResNet encoder coupled to a Gaussian neural density estimator can be pretrained, before fine-tuning on the available real simulations.
{Training the same architecture from scratch on identical simulation subsets remains worse on average with all $672$ real fields than the pretrained model with none. Synthetic pretraining improves the validation NLL by $0.76$ (from $-0.93$ to $-1.69$) before \emph{any} genuine simulation is used for training} (Figure~\ref{fig:sbi-summary}).
This shows that macrocanonical surrogates can unblock amortized scientific pipelines whose limiting factor is simulator throughput rather than model capacity.
Appendix~\ref{app:throughput-sbi} gives further details.

\paragraph{Inductive-bias ablation.}
\label{sec:ablation}
Finally we isolate the architectural biases advocated in Sections~\ref{sec:residual}--\ref{sec:entropy}: identity residual initialization so Jacobian spectra begin at volume-preserving values, learning rates mild enough for Theorem~\ref{thm:preconditioned}'s linearization, and circular convolutions enforcing translation equivariance so Corollary~\ref{cor:stationarity} applies on the torus.
Removing each ingredient (training at $\eta=10^{-2}$, deleting residual shortcuts, or swapping zero padding and circular padding) while freezing datasets and iteration budgets reproduces the predicted failure modes: collapsed transports, rapid entropy loss, or stationary-boundary artifacts (Figure~\ref{fig:ablation-inductive-bias}).

\section{Discussion and Limitations}
\vspace{-0.3cm}
\label{sec:limitations}

We have introduced macrocanonical generator networks, an amortized neural surrogate for stationary physical processes that matches multiscale feature statistics in expectation, together with the entropy- and dynamics-based analysis (Sections~\ref{sec:dynamics}--\ref{sec:entropy}) that identifies identity-initialized residual parameterizations and small learning rates as the inductive biases responsible for diversity-preserving training.
On the cosmological dark-energy inference task of Section~\ref{sec:sbi} (a real-world scientific task whose limiting factor is $N$-body simulator throughput)  pretraining with the macrocanonical-generated surrogates makes simulation-based inference of dark energy from matter-density image data tractable in a regime where the available real-simulation budget alone is insufficient. 

\vspace{-0.1cm}
Limitation discussion:
\vspace{-0.25cm}

\begin{itemize}[leftmargin=.9em, rightmargin=0pt]

\item The method depends on the choice of feature statistics.
If the chosen statistics do not capture the relevant structure of the target process, generated samples may fail to reproduce important properties.
\vspace{-0.1cm}

\item The framework is designed for stationary processes with periodic boundary conditions, where translation invariance provides the symmetry group that enables estimation from few realizations.
For non-stationary processes the method cannot directly exploit this symmetry, and the data efficiency advantages may not apply.
An interesting direction for future work is to pretrain generators on stationary processes (e.g., periodic fluid simulations) and then fine-tune on non-stationary simulations, which we have not yet tested.
{The periodic-boundary requirement can be relaxed, for example by appropriately masking or padding the region near the boundary (naive non-periodic application of the method instead produces the boundary artifacts in Figure~\ref{fig:ablation-inductive-bias}).}
\vspace{-0.1cm}

\item While the generator often appears empirically to produce samples consistent with maximum-entropy distributions under the imposed constraints, and we provide theoretical justification for this, we cannot currently provide a general theoretical guarantee of convergence to such a distribution (similarly, neither do \cite{bruna2018multiscale,zhang2021maximum}){, nor of avoiding collapse (Section~\ref{sec:entropy})}.

\end{itemize}

\begin{ack}
NJ thanks Benjamin Joachimi and Lorne Whiteway for useful discussions. The authors acknowledge support from the Simons Collaboration on ``Learning the Universe''. NJ acknowledges support from the King's AI+ Academic Fellowship at King's College London.
\end{ack}

\bibliographystyle{plainnat}
\bibliography{references}

\begin{thebibliography}{35}
\providecommand{\natexlab}[1]{#1}
\providecommand{\url}[1]{\texttt{#1}}
\expandafter\ifx\csname urlstyle\endcsname\relax
  \providecommand{\doi}[1]{doi: #1}\else
  \providecommand{\doi}{doi: \begingroup \urlstyle{rm}\Url}\fi

\bibitem[Allys et~al.(2020)Allys, Marchand, Cardoso, Villaescusa-Navarro, Ho,
  and Mallat]{allys2020new}
Erwan Allys, Th{\'e}o Marchand, Jean-Fran{\c{c}}ois Cardoso, Francisco
  Villaescusa-Navarro, Shirley Ho, and St{\'e}phane Mallat.
\newblock New interpretable statistics for large-scale structure analysis and
  generation.
\newblock \emph{Physical Review D}, 102\penalty0 (10):\penalty0 103506, 2020.

\bibitem[Alsing et~al.(2019)Alsing, Charnock, Feeney, and
  Wandelt]{alsing2019fast}
Justin Alsing, Tom Charnock, Stephen Feeney, and Benjamin Wandelt.
\newblock Fast likelihood-free cosmology with neural density estimators and
  active learning.
\newblock \emph{Monthly Notices of the Royal Astronomical Society}, 2019.
\newblock \doi{10.1093/mnras/stz1960}.

\bibitem[Behrens and Dias(2015)]{behrens2015tsunami}
J{\"o}rn Behrens and Fr{\'e}d{\'e}ric Dias.
\newblock New computational methods in tsunami science.
\newblock \emph{Philosophical Transactions of the Royal Society A},
  373\penalty0 (2053):\penalty0 20140382, 2015.

\bibitem[Bruna and Mallat(2018)]{bruna2018multiscale}
Joan Bruna and St{\'e}phane Mallat.
\newblock Multiscale sparse microcanonical models.
\newblock \emph{Mathematical Statistics and Learning}, 1:\penalty0 257--315,
  2018.

\bibitem[Cheng and M{\'e}nard(2021)]{cheng2024scattering}
Sihao Cheng and Brice M{\'e}nard.
\newblock Weak lensing scattering transform: dark energy and neutrino mass
  sensitivity.
\newblock \emph{Monthly Notices of the Royal Astronomical Society},
  507\penalty0 (1):\penalty0 1012--1020, 2021.

\bibitem[Cover and Thomas(2006)]{cover2006elements}
Thomas~M Cover and Joy~A Thomas.
\newblock \emph{Elements of Information Theory}.
\newblock Wiley-Interscience, 2006.

\bibitem[Cranmer et~al.(2020)Cranmer, Brehmer, and Louppe]{cranmer2020frontier}
Kyle Cranmer, Johann Brehmer, and Gilles Louppe.
\newblock The frontier of simulation-based inference.
\newblock \emph{Proceedings of the National Academy of Sciences}, 117\penalty0
  (48):\penalty0 30055--30062, 2020.

\bibitem[Favero et~al.(2025)Favero, Sclocchi, Cagnetta, Frossard, and
  Wyart]{favero2025compositional}
Alessandro Favero, Antonio Sclocchi, Francesco Cagnetta, Pascal Frossard, and
  Matthieu Wyart.
\newblock How compositional generalization and creativity improve as diffusion
  models are trained.
\newblock In \emph{Proceedings of the 42nd International Conference on Machine
  Learning}, 2025.

\bibitem[Goodfellow et~al.(2014)Goodfellow, Pouget-Abadie, Mirza, Xu,
  Warde-Farley, Ozair, Courville, and Bengio]{goodfellow2014generative}
Ian Goodfellow, Jean Pouget-Abadie, Mehdi Mirza, Bing Xu, David Warde-Farley,
  Sherjil Ozair, Aaron Courville, and Yoshua Bengio.
\newblock Generative adversarial nets.
\newblock In \emph{Advances in Neural Information Processing Systems}, pages
  2672--2680, 2014.

\bibitem[Gulrajani et~al.(2017)Gulrajani, Ahmed, Arjovsky, Dumoulin, and
  Courville]{gulrajani2017improved}
Ishaan Gulrajani, Faruk Ahmed, Martin Arjovsky, Vincent Dumoulin, and Aaron
  Courville.
\newblock Improved training of {W}asserstein {GAN}s.
\newblock In \emph{Advances in Neural Information Processing Systems},
  volume~30, 2017.

\bibitem[He et~al.(2016)He, Zhang, Ren, and Sun]{he2016deep}
Kaiming He, Xiangyu Zhang, Shaoqing Ren, and Jian Sun.
\newblock Deep residual learning for image recognition.
\newblock In \emph{Proceedings of the IEEE Conference on Computer Vision and
  Pattern Recognition}, pages 770--778, 2016.

\bibitem[Hikida et~al.(2025)Hikida, Bharti, Jeffrey, and
  Briol]{hikida2025multilevel}
Yuga Hikida, Ayush Bharti, Niall Jeffrey, and Fran{\c{c}}ois-Xavier Briol.
\newblock Multilevel neural simulation-based inference.
\newblock \emph{Advances in Neural Information Processing Systems}, 38, 2025.

\bibitem[Ho et~al.(2020)Ho, Jain, and Abbeel]{ho2020denoising}
Jonathan Ho, Ajay Jain, and Pieter Abbeel.
\newblock Denoising diffusion probabilistic models.
\newblock In \emph{Advances in Neural Information Processing Systems},
  volume~33, pages 6840--6851, 2020.

\bibitem[Hoppe et~al.(2021)Hoppe, Embreus, and F{\"u}l{\"o}p]{hoppe2021dream}
Mathias Hoppe, Ola Embreus, and T{\"u}nde F{\"u}l{\"o}p.
\newblock {DREAM}: A fluid-kinetic framework for tokamak disruption runaway
  electron simulations.
\newblock \emph{Computer Physics Communications}, 268:\penalty0 108098, 2021.

\bibitem[Jacot et~al.(2018)Jacot, Gabriel, and Hongler]{jacot2018neural}
Arthur Jacot, Franck Gabriel, and Cl{\'e}ment Hongler.
\newblock Neural tangent kernel: Convergence and generalization in neural
  networks.
\newblock In \emph{Advances in Neural Information Processing Systems},
  volume~31, 2018.

\bibitem[Jaynes(1957)]{jaynes1957information}
Edwin~T Jaynes.
\newblock Information theory and statistical mechanics.
\newblock \emph{Physical Review}, 106\penalty0 (4):\penalty0 620, 1957.

\bibitem[Jeffrey et~al.(2025)]{jeffrey2025dark}
Niall Jeffrey et~al.
\newblock Dark energy survey year 3 results: likelihood-free, simulation-based
  {wCDM} inference with neural compression of weak-lensing map statistics.
\newblock \emph{Monthly Notices of the Royal Astronomical Society},
  536\penalty0 (2):\penalty0 1303--1322, 2025.

\bibitem[Kirby et~al.(2023)Kirby, Briol, Dunstan, and Nishino]{kirby2023wind}
Andrew Kirby, Fran{\c{c}}ois-Xavier Briol, Thomas~D Dunstan, and Takafumi
  Nishino.
\newblock Data-driven modelling of turbine wake interactions and flow
  resistance in large wind farms.
\newblock \emph{Wind Energy}, 26\penalty0 (9):\penalty0 875--1011, 2023.

\bibitem[Lempereur et~al.(2026)Lempereur, Cuvelle-Magar, Coeurdoux, Mallat, and
  Vanden-Eijnden]{lempereur2026mgd}
Etienne Lempereur, Nathanael Cuvelle-Magar, Florentin Coeurdoux, Stephane
  Mallat, and Eric Vanden-Eijnden.
\newblock {MGD}: Moment guided diffusion for maximum entropy generation.
\newblock \emph{arXiv preprint arXiv:2602.17211}, 2026.

\bibitem[Li et~al.(2017)Li, Chang, Cheng, Yang, and P{\'o}czos]{li2017mmdgan}
Chun-Liang Li, Wei-Cheng Chang, Yu~Cheng, Yiming Yang, and Barnab{\'a}s
  P{\'o}czos.
\newblock {MMD GAN}: Towards deeper understanding of moment matching network.
\newblock In \emph{Advances in Neural Information Processing Systems},
  volume~30, 2017.

\bibitem[Li et~al.(2015)Li, Swersky, and Zemel]{li2015gmmn}
Yujia Li, Kevin Swersky, and Richard Zemel.
\newblock Generative moment matching networks.
\newblock In \emph{International Conference on Machine Learning (ICML)}, 2015.

\bibitem[Lipman et~al.(2023)Lipman, Chen, Ben-Hamu, Nickel, and
  Le]{lipman2023flow}
Yaron Lipman, Ricky T.~Q. Chen, Heli Ben-Hamu, Maximilian Nickel, and Matthew
  Le.
\newblock Flow matching for generative modeling.
\newblock In \emph{International Conference on Learning Representations}, 2023.

\bibitem[Mallat et~al.(2020)Mallat, Zhang, and Rochette]{mallat2020phase}
St{\'e}phane Mallat, Sixin Zhang, and Gaspar Rochette.
\newblock Phase harmonic correlations and convolutional neural networks.
\newblock \emph{Information and Inference: A Journal of the IMA}, 9\penalty0
  (3):\penalty0 721--747, 2020.

\bibitem[McCabe et~al.(2025)McCabe, Mukhopadhyay, Marwah,
  Regaldo-Saint~Blancard, Rozet, Diaconu, Meyer, Wong, Sotoudeh, Bietti,
  et~al.]{mccabe2025walrus}
Michael McCabe, Payel Mukhopadhyay, Tanya Marwah, Bruno Regaldo-Saint~Blancard,
  Francois Rozet, Cristiana Diaconu, Lucas Meyer, Kaze W.~K. Wong, Hadi
  Sotoudeh, Alberto Bietti, et~al.
\newblock Walrus: A cross-domain foundation model for continuum dynamics.
\newblock \emph{arXiv preprint arXiv:2511.15684}, 2025.

\bibitem[McFadden(1989)]{mcfadden1989msm}
Daniel McFadden.
\newblock A method of simulated moments for estimation of discrete response
  models without numerical integration.
\newblock \emph{Econometrica}, 57\penalty0 (5):\penalty0 995--1026, 1989.

\bibitem[{Ohana} et~al.(2024){Ohana}, {McCabe}, {Meyer}, {Morel}, {Agocs},
  {Beneitez}, {Berger}, {Burkhart}, {Burns}, {Dalziel}, {Fielding},
  {Fortunato}, {Goldberg}, {Hirashima}, {Jiang}, {Kerswell}, {Maddu}, {Miller},
  {Mukhopadhyay}, {Nixon}, {Shen}, {Watteaux}, {R{\'e}galdo-Saint Blancard},
  {Rozet}, {Parker}, {Cranmer}, and {Ho}]{ohana2024well}
Ruben {Ohana}, Michael {McCabe}, Lucas {Meyer}, Rudy {Morel}, Fruzsina~J.
  {Agocs}, Miguel {Beneitez}, Marsha {Berger}, Blakesley {Burkhart}, Keaton
  {Burns}, Stuart~B. {Dalziel}, Drummond~B. {Fielding}, Daniel {Fortunato},
  Jared~A. {Goldberg}, Keiya {Hirashima}, Yan-Fei {Jiang}, Rich~R. {Kerswell},
  Suryanarayana {Maddu}, Jonah {Miller}, Payel {Mukhopadhyay}, Stefan~S.
  {Nixon}, Jeff {Shen}, Romain {Watteaux}, Bruno {R{\'e}galdo-Saint Blancard},
  Fran{\c{c}}ois {Rozet}, Liam~H. {Parker}, Miles {Cranmer}, and Shirley {Ho}.
\newblock {The Well: a Large-Scale Collection of Diverse Physics Simulations
  for Machine Learning}.
\newblock \emph{arXiv e-prints}, art. arXiv:2412.00568, November 2024.
\newblock \doi{10.48550/arXiv.2412.00568}.

\bibitem[Parker et~al.(2025)Parker, Lanusse, Shen, Liu, Hehir, Sarra, Meyer,
  Bowles, Wagner-Carena, Qu, Golkar, Bietti, et~al.]{parker2025aion}
Liam Parker, Francois Lanusse, Jeff Shen, Ollie Liu, Tom Hehir, Leopoldo Sarra,
  Lucas Meyer, Micah Bowles, Sebastian Wagner-Carena, Helen Qu, Siavash Golkar,
  Alberto Bietti, et~al.
\newblock {AION-1}: Omnimodal foundation model for astronomical sciences.
\newblock \emph{arXiv preprint arXiv:2510.17960}, 2025.

\bibitem[Pope(2000)]{pope2000turbulent}
Stephen~B. Pope.
\newblock \emph{Turbulent Flows}.
\newblock Cambridge University Press, 2000.

\bibitem[Portilla and Simoncelli(2000)]{portilla2000parametric}
Javier Portilla and Eero~P Simoncelli.
\newblock A parametric texture model based on joint statistics of complex
  wavelet coefficients.
\newblock \emph{International Journal of Computer Vision}, 40\penalty0
  (1):\penalty0 49--70, 2000.

\bibitem[Regaldo-Saint~Blancard et~al.(2020)Regaldo-Saint~Blancard, Levrier,
  Allys, Bellomi, and Boulanger]{regaldo2020rwst}
B.~Regaldo-Saint~Blancard, F.~Levrier, E.~Allys, E.~Bellomi, and F.~Boulanger.
\newblock Statistical description of dust polarized emission from the diffuse
  interstellar medium: A rwst approach.
\newblock \emph{Astronomy \& Astrophysics}, 642:\penalty0 A217, October 2020.
\newblock ISSN 1432-0746.
\newblock \doi{10.1051/0004-6361/202038044}.
\newblock URL \url{http://dx.doi.org/10.1051/0004-6361/202038044}.

\bibitem[Rezende and Mohamed(2015)]{rezende2015variational}
Danilo Rezende and Shakir Mohamed.
\newblock Variational inference with normalizing flows.
\newblock In \emph{International Conference on Machine Learning}, pages
  1530--1538, 2015.

\bibitem[Villaescusa-Navarro et~al.(2020)Villaescusa-Navarro, Hahn, Massara,
  Banerjee, Delgado, Ramanah, Charnock, Giusarma, Li, Allys,
  et~al.]{villaescusa2020quijote}
Francisco Villaescusa-Navarro, ChangHoon Hahn, Elena Massara, Arka Banerjee,
  Andrij~M. Delgado, Doogesh~Kodi Ramanah, Tom Charnock, Elena Giusarma, Yin
  Li, Erwan Allys, et~al.
\newblock The {Quijote} simulations.
\newblock \emph{The Astrophysical Journal Supplement Series}, 250\penalty0
  (1):\penalty0 2, 2020.

\bibitem[Villaescusa-Navarro et~al.(2023)]{villaescusa2023camels}
Francisco Villaescusa-Navarro et~al.
\newblock The {CAMELS} project: Public data release.
\newblock \emph{The Astrophysical Journal Supplement Series}, 265\penalty0
  (2):\penalty0 54, 2023.

\bibitem[Wiesner et~al.(2025)Wiesner, Wessling, and Baek]{wiesner2025physics}
Florian Wiesner, Matthias Wessling, and Stephen Baek.
\newblock Towards a physics foundation model.
\newblock \emph{arXiv preprint arXiv:2509.13805}, 2025.

\bibitem[Zhang and Mallat(2021)]{zhang2021maximum}
Sixin Zhang and St{\'e}phane Mallat.
\newblock Maximum entropy models from phase harmonic covariances.
\newblock \emph{Applied and Computational Harmonic Analysis}, 53:\penalty0
  199--230, 2021.

\end{thebibliography}


\appendix

\section{Proof of Theorem~\ref{thm:preconditioned}}
\vspace{-0.3cm}
\label{app:proof-thm1}

We give the complete derivation of the kernel-preconditioned sample-space update stated in Theorem~\ref{thm:preconditioned}; a high-level argument is sketched in the main text.

\begin{proof}[Proof of Theorem~\ref{thm:preconditioned}]
Let
\begin{equation}
m(\theta)=\mathbb{E}_{x' \sim p_0}\bigl[\phi(F_\theta(x'))\bigr],
\qquad
L(\theta)=\frac12\|m(\theta)-\phi^\ast\|^2.
\end{equation}
A gradient descent step gives
\begin{equation}
\theta^{(t+1)}=\theta^{(t)}-\eta \nabla_\theta L(\theta^{(t)}).
\end{equation}
Because the learning rate is assumed to be sufficiently small, the parameter update is small, so for a fixed input $x$ we may linearize the generator around $\theta^{(t)}${\ , with remainder $O(\|\Delta\theta\|^{2})$ by assumption (i)}:
\begin{equation}
F_{\theta^{(t)}+\Delta\theta}(x)
\approx
F_{\theta^{(t)}}(x)+J_{\theta^{(t)}}(x)\Delta\theta,
\end{equation}
where
\begin{equation}
J_{\theta^{(t)}}(x)=\frac{\partial F_{\theta^{(t)}}(x)}{\partial \theta}.
\end{equation}
Writing $y^{(t)}(x)=F_{\theta^{(t)}}(x)$ and $\Delta\theta=-\eta \nabla_\theta L(\theta^{(t)})$, this gives
\begin{equation}
y^{(t+1)}(x)
\approx
y^{(t)}(x)-\eta J_{\theta^{(t)}}(x)\nabla_\theta L(\theta^{(t)}).
\end{equation}

It remains to compute $\nabla_\theta L$.
By the chain rule,
\begin{equation}
\nabla_\theta L(\theta)
=
\left(\frac{\partial m(\theta)}{\partial \theta}\right)^\top
\bigl(m(\theta)-\phi^\ast\bigr).
\end{equation}
Assuming differentiation may be interchanged with expectation,
\begin{equation}
\frac{\partial m(\theta)}{\partial \theta}
=
\mathbb{E}_{x' \sim p_0}
\left[
\frac{\partial \phi(F_\theta(x'))}{\partial \theta}
\right].
\end{equation}
For each sample $x'$, the chain rule gives
\begin{equation}
\frac{\partial \phi(F_\theta(x'))}{\partial \theta}
=
\nabla_y \phi(F_\theta(x'))\, J_\theta(x').
\end{equation}
Define
\begin{equation}
\nabla_y \widetilde L_{x'}
=
\left(\nabla_y \phi(F_\theta(x'))\right)^\top
\bigl(m(\theta)-\phi^\ast\bigr).
\end{equation}
Then
\begin{equation}
\nabla_\theta L(\theta)
=
\mathbb{E}_{x' \sim p_0}
\left[
J_\theta(x')^\top \nabla_y \widetilde L_{x'}
\right].
\end{equation}
Substituting into the linearized sample update yields
\begin{equation}
y^{(t+1)}(x)
\approx
y^{(t)}(x)
-\eta
\mathbb{E}_{x' \sim p_0}
\left[
J_{\theta^{(t)}}(x)J_{\theta^{(t)}}(x')^\top
\nabla_y \widetilde L_{x'}
\right].
\end{equation}
Defining
\begin{equation}
K_{\theta^{(t)}}(x,x')
=
J_{\theta^{(t)}}(x)J_{\theta^{(t)}}(x')^\top
\end{equation}
gives the stated result.
\end{proof}

\section{Proof of Corollary~\ref{cor:micro-inheritance}}
\vspace{-0.3cm}
\label{app:proof-cor2}

\begin{proof}[Proof of Corollary~\ref{cor:micro-inheritance}]
By assumption $\phi$ is shift-invariant, so for any translation $g$ and input $z$, $\phi(gz) = \phi(z)$, and the residual $\phi(F_\theta(x'))-\phi^\ast$ is invariant under $x' \mapsto gx'$. Periodic translations act by permutation matrices, so $g^\top = g^{-1}$. Differentiating $\phi(gy)=\phi(y)$ in $y$ and using translation equivariance of $F_\theta$ gives $\nabla_y\widetilde L_{gx'} = g\,\nabla_y\widetilde L_{x'}$. Translation equivariance of $F_\theta$ also gives $J_\theta(gx) = g\,J_\theta(x)$, hence $K_\theta(gx, gx') = g\,K_\theta(x,x')\,g^{-1}$. The integrand of the per-sample update of Theorem~\ref{thm:preconditioned} therefore satisfies
\begin{equation}
K_\theta(gx, gx')\,\nabla_y\widetilde L_{gx'} = g\bigl(K_\theta(x,x')\,\nabla_y\widetilde L_{x'}\bigr).
\end{equation}
Since $p_0$ is invariant under $g$, the change of variables $x' \mapsto gx'$ leaves the expectation unchanged, so the full update map $y^{(t)}(x) \mapsto y^{(t+1)}(x)$ is shift-equivariant.
Stationarity of the {\ generated distribution} of $y^{(t)}$ at every {\ training step} $t$ then follows from Proposition~\ref{prop:symmetry} applied to $F_{\theta^{(t)}}$, which is translation-equivariant for every $\theta^{(t)}$.
\end{proof}

\section{Quijote $N$-body simulations and slice preprocessing}
\vspace{-0.3cm}
\label{app:quijote}

The cosmic-web experiments of Section~\ref{sec:cosmo} use 2-D slices extracted from the Quijote suite of $N$-body simulations \cite{villaescusa2020quijote}, a publicly released ensemble of $\mathcal{O}(10^4)$ dark-matter-only simulations spanning a wide range of cosmological parameters and designed to support large-scale-structure cosmology and machine-learning applications.
The SBI experiment of Section~\ref{sec:sbi} uses the same suite but draws cosmologies from the Quijote Latin-hypercube subset rather than the single-cosmology fiducial pool described below; its preprocessing differs only in extracting non-overlapping mean-projections of voxel thickness $4$ along $z$ to produce $32$ images per cosmology.

\paragraph{Simulation setup.}
Each Quijote run evolves a set of dark-matter particles inside a periodic cubic box from a high-redshift initial condition --- generated by second-order Lagrangian perturbation theory from a linear matter power spectrum --- down to redshift $z = 0$ using a particle-based gravity solver.
After the gravitational evolution, the particles are deposited onto a regular three-dimensional density grid using a mass-assignment scheme.
For our experiments we use the \emph{fiducial} cosmology, in which the cosmological parameters are held fixed at values consistent with current cosmic-microwave-background and large-scale-structure constraints; this single-cosmology subset provides a large pool of statistically independent realizations of the matter density field, all governed by the same underlying physics.

\paragraph{Slice extraction.}
From each three-dimensional density snapshot we take a 2-D slice through the volume at $z = 0$, producing a square map of the projected matter density.
We take $100$ such slices in total, drawn from the fiducial Quijote pool, and treat them as independent realizations of the same stationary 2-D random field with periodic boundary conditions inherited from the simulation box.

\paragraph{Preprocessing.}
The raw surface-density values span several orders of magnitude due to the highly non-Gaussian, filamentary structure of the cosmic web.
To compress this dynamic range we apply $x \mapsto \log_{10}(x + 1)$ to the surface density of each slice.
We then downsample to a $128 \times 128$ pixel grid by spatial mean-pooling.
The pixel values are subsequently normalized to zero mean and unit variance using the empirical mean and standard deviation computed over the entire pool of slices, and finally rescaled by a constant factor of $1/8$ so that the resulting fields lie comfortably inside the dynamic range of the WPH operator.
The resulting slices are stationary, periodic on the simulation grid, and exhibit the characteristic non-Gaussian filamentary structure that motivates the use of higher-order summary statistics such as wavelet phase harmonic coefficients.

\section{Throughput and simulation-based inference protocol}
\label{app:throughput-sbi}

\paragraph{Reference $N$-body cost.}
A single $N$-body simulation of the kind used in modern weak-lensing cosmology, for example the Gower Street suite produced for recent Dark Energy Survey analyses~\cite{jeffrey2025dark}, typically requires $\mathcal{O}(\text{hours to days})$ of compute per realization, even when distributed across hundreds of cores.
The Quijote suite required more than $35$ million CPU core-hours for its $44{,}100$ simulations, about $800$ core-hours per realization~\cite{villaescusa2020quijote}, so even $30$ Latin-hypercube cosmologies constitute an aggressive pool for amortized inference.

\paragraph{Macrocanonical inference throughput.}
Once trained, the macrocanonical generator produces a batch of $200$ new $128{\times}128$ realizations in one forward pass that completes in well under a second on a single NVIDIA A100 GPU---a few milliseconds per field and more than six orders of magnitude higher throughput than a single Gower-Street-class $N$-body realization.

\paragraph{Training wall-clock.}
Training uses two A100 GPUs at batch size $200$, costing roughly one minute per $25$ iterations.
The simulation-based inference configurations use $500$ iterations ($\sim$$20$ minutes); the higher-fidelity runs for Section~\ref{sec:cosmo} use $5000$ iterations (a few hours).

\paragraph{Microcanonical synthesis baseline.}
Microcanonical gradient-descent synthesis~\cite{bruna2018multiscale,zhang2021maximum} solves a fresh inner loop per sample and takes approximately $20$~seconds per $128{\times}128$ field on the same A100---more than $4{\times}10^3$ slower per sample than the amortized generator---while pinning $\phi=\phi^\ast$ exactly (Section~\ref{sec:cosmo}), so it cannot supply the fluctuations downstream inference requires.

\paragraph{Amortization.}
One-off generator training is amortized across all later queries, making simulation-based inference, uncertainty quantification, and sensitivity analysis feasible where repeated $N$-body runs are not.

\paragraph{SBI data splits and corruption.}
We draw $32$ non-overlapping $128 \times 128$ projections from each of the first $30$ Quijote Latin-hypercube simulations, split by cosmology into $24$ training cosmologies ($768$ training fields) and $6$ validation cosmologies ($192$ validation fields).
The same fixed instrument-footprint corruption (heteroskedastic noise with per-pixel standard-deviation map, jagged border, centred-circle mask) is applied to every real and synthetic image.
We use a small ResNet embedding $f_\psi$ and a single-component Gaussian estimator $p(\Omega_{\rm DE} \mid \mathrm{image}) = \mathcal{N}(\mu(f_\psi), \sigma(f_\psi)^2)$, with Adam at $5{\times}10^{-5}$ for fine-tuning and $5{\times}10^{-4}$ for from-scratch training, plus patience-based early stopping.

\paragraph{Pretraining and fine-tuning.}
We train one macrocanonical generator per training cosmology from a single $128 \times 128$ WPH target, draw $1000$ synthetic fields per cosmology, yielding $24{,}000$ synthetic training fields compared with $768$ real training tiles ($24\times 32$ non-overlapping projections).
We pretrain the encoder and Gaussian neural density estimator on synthetics, then fine-tune on real subsets of size {\ $N \in \{1, 10, 100, 300, 672\}$ ($96$ of the $768$ training tiles, four projections per training cosmology, are held out for early stopping of all models)}.
An identical architecture trained from scratch on the same real subsets is the baseline.

\paragraph{Likelihood results.}
{\ Validation NLL is averaged over the four validation cosmologies whose $\Omega_{\rm DE}$ lies inside the training range $[0.525, 0.887]$ (the other two require extrapolation) and over three seeds.
Training from scratch stays at the empirical prior ($-0.93$) up to $N{=}10$ and reaches $-1.4 \pm 0.3$ at $N \geq 300$; pretraining alone gives $-1.69 \pm 0.03$ before any genuine simulation is used for training, and fine-tuning reaches $-1.76$ at $N{=}672$} (Figure~\ref{fig:sbi-summary}).
\paragraph{Synthetic ensemble cost.}
After $\sim$$20$ minutes of generator training per cosmology on two A100s, each network emits $\sim$$200$ fields per second.
The resulting $24{,}000$-tile synthetic training corpus would correspond to $750$ additional Quijote realizations, roughly $6\times10^{5}$ CPU core-hours, for comparably dense cosmology--image supervision.

\section{Experimental setup details}
\vspace{-0.3cm}
\label{app:experiment-details}

This appendix gives the architecture, initialization, and optimizer details summarized in Section~\ref{sec:setup}.

{\ \paragraph{Code availability.}
A reference implementation of the macrocanonical generator, with training and sampling scripts for both benchmarks of Section~\ref{sec:experiments}, is available at \url{https://github.com/NiallJeffrey/MacrocanonicalGeneratorNetwork}.}

\paragraph{Generator architectures.}
We instantiate the two residual parameterizations introduced in Section~\ref{sec:macro}.
The \emph{additive} form of eq.~\eqref{eq:additive} uses a single convolutional U-Net for $G_\theta$.
The \emph{multiplicative} form of eq.~\eqref{eq:multiplicative} uses two independent U-Nets of identical architecture for $G_\theta$ and $H_\theta$.
Both networks share the same backbone: a four-stage encoder--decoder with channel widths $64\!\to\!1024$, $3{\times}3$ convolutions, batch normalization, and ReLU activations.
Because every target process in our experiments is periodic, all convolutions use circular padding and all bilinear upsamples are wrapped in a circular pad/interpolate/crop block, so the network is exactly translation-equivariant on the periodic grid (Corollary~\ref{cor:stationarity}).
Convolutional weights are initialized from $\mathcal{N}(0, 10^{-4})$ so that $G_{\theta_0}(x) \approx 0$; for the multiplicative parameterization, an additive constant is appended to the forward map of $H_\theta$ so that $H_{\theta_0}(x) \approx \mathbf{1}$.
Both parameterizations therefore begin training at the identity, $F_{\theta_0}(x) = x$, as required by the analysis of Sections~\ref{sec:dynamics} and~\ref{sec:entropy}.
Activation checkpointing is used in the U-Net body to reduce peak memory at the cost of an additional forward pass during the backward, allowing batch size $200$ at $128{\times}128$ resolution on a single GPU.

\paragraph{Feature statistics.}
We use wavelet phase harmonic (WPH) coefficients~\cite{mallat2020phase,zhang2021maximum} as the feature map $\phi$, computed under periodic boundary conditions.
The hyperparameters $(J, L, \Delta\ell, \Delta n)$ and the phase-harmonic exponent list are chosen per dataset; the per-benchmark values used in this work are summarized in Appendix~\ref{app:wph}, Table~\ref{tab:wph-config}.
The training target $\phi^\ast$ is the empirical mean of the WPH coefficients over the available training realizations; in the single-realization regime this reduces to the WPH coefficients of that one signal.

{\ \paragraph{MHD benchmark data.}
The single $128{\times}128$ MHD turbulence snapshot of Section~\ref{sec:experiments} is derived from a $512{\times}512$ map of simulated dust emission from the magnetohydrodynamic simulation of the diffuse interstellar medium used by \citet{regaldo2020rwst}.
The map is stored as a complex-valued array; we take its modulus, apply a $\log_{10}$ transform, mean-pool to $128{\times}128$, subtract a smooth large-scale component (a Gaussian filter of width $8$ pixels with periodic wrapping), standardize to zero mean and unit variance, and rescale by $1/16$.}

\paragraph{Input distribution.}
$p_0$ is a stationary Gaussian random field with periodic boundary conditions, generated by FFT from the radially averaged power spectrum estimated from the training data.
This matches the second-order statistics of the target process while leaving all higher-order structure to be learned by $F_\theta$.

\paragraph{Optimization.}
We minimize the macrocanonical loss with Adam at base learning rate $10^{-4}$ and global gradient-norm clipping at $1.0$, optionally preceded by a short linear warmup and followed by cosine annealing over the configured number of iterations.
The generator and the WPH operator are placed on separate GPUs to accommodate their respective memory footprints; cross-device autograd is handled transparently by PyTorch.

{\ \section{MHD turbulence suite for the controlled comparisons}
\label{app:mhd64}

The training-set-size sweep and the baseline comparison of Section~\ref{sec:experiments}, and the sweeps of Appendix~\ref{app:sweeps}, require many independent realizations of one stationary process: to train at $N_{\text{train}}=10$ and $100$, and to score every method against held-out fields that no method has seen.
The single MHD snapshot of the main benchmark cannot supply these, so for these experiments we use the \texttt{MHD\_64} dataset of The Well \cite{ohana2024well}: three-dimensional isothermal, compressible MHD turbulence without self-gravity, simulated at $256^3$ with periodic boundaries and released at $64^3$ after an ideal low-pass filter, with independent runs for each combination of sonic and Alfv\'enic Mach number.
We use the single combination $\mathcal{M}_s=2$, $\mathcal{M}_A=0.7$, so that all fields are realizations of one stationary process, and take two-dimensional $64{\times}64$ slices of the density at fixed $z$.
The low-pass filter can produce small non-positive density values, which we clamp to the smallest positive value before taking $\log_{10}$.
Training fields are drawn from the first of the $8$ runs in the released training file, subsampled every $8$ slices in $z$ and every $5$ snapshots in time to reduce correlation; the $100$ held-out evaluation fields are drawn in the same way from the remaining $7$ runs.
The held-out fields are therefore independent simulations, whereas the $10$ or $100$ training fields are correlated slices of a single run.
Fields are standardized with the mean and standard deviation of the training fields only and divided by $8$; the WPH configuration is $J=4$, $L=4$, $\Delta n=1$, $\Delta\ell=2$ with the default phase-harmonic exponents.
Every method in Tables~\ref{tab:nsweep} and~\ref{tab:baselines} is trained and scored on exactly these fields.

The macrocanonical generator uses the multiplicative parameterization~\eqref{eq:multiplicative} with the optimizer settings of Appendix~\ref{app:experiment-details}, trained for $1500$ iterations at batch size $64$ and without augmentation.
The learned baselines are trained for $20{,}000$ iterations at batch size $32$: WGAN-GP \cite{gulrajani2017improved} with five critic steps per generator step and gradient-penalty weight $10$; flow matching \cite{lipman2023flow} and the DDPM \cite{ho2020denoising} with the same $5.3$M-parameter circular-padded U-Net, sampled with $100$ Euler steps and $250$ DDIM steps respectively.
During DDPM sampling we clamp the predicted clean field to the range of the training data (its $0.1$st to $99.9$th percentiles, widened by $10\%$); without this safeguard the sampler diverges on these fields.
Translation augmentation, where used, applies a random cyclic shift to the training field at every step.
Microcanonical synthesis runs $400$ steps of pixel-space gradient descent per sample.}

{\ \paragraph{Metrics.}
The WPH distance is the root-mean-square difference, over the normalized WPH coefficients, between the mean feature vector of the generated samples and that of the held-out fields.
KS mean-$p$ is the mean, over WPH coefficients, of the two-sample Kolmogorov--Smirnov $p$-value comparing the across-sample distribution of that coefficient for generated and held-out fields: it is near $0.5$ when the per-realization distributions match and near $0$ when the scatter is missing or shifted.
The nearest-neighbour (NN) ratio is the root-mean-square distance from a sample to its nearest training field, minimized over cyclic translations, divided by the same quantity for held-out fields; values much less than $1$ indicate memorization, including memorization of shifted copies.
Numbers in parentheses are held-out fields scored against themselves, which is the level a faithful surrogate should reach.

\begin{table}[h]
  \centering
  \small
  \caption{Effect of training-set size ($64{\times}64$ MHD suite of Appendix~\ref{app:mhd64}, scored against $100$ held-out fields; parentheses: held-out fields scored against themselves).}
  \label{tab:nsweep}
  \begin{tabular}{lccc}
    \toprule
    $N_{\text{train}}$ & WPH distance & KS mean-$p$ & NN ratio \\
    \midrule
    $1$   & $0.20$ ($0.053$)  & $0.08$ ($0.54$) & $0.89$ ($1.00$) \\
    $10$  & $0.075$ ($0.056$) & $0.27$ ($0.47$) & $0.98$ ($0.96$) \\
    $100$ & $0.086$ ($0.037$) & $0.31$ ($0.50$) & $0.99$ ($1.01$) \\
    \bottomrule
  \end{tabular}
\end{table}

\begin{table}[h]
  \centering
  \small
  \caption{Baseline comparison on the $64{\times}64$ MHD suite of Appendix~\ref{app:mhd64}: all methods trained on the same fields and scored against $100$ held-out fields. NN ratios much less than $1$ mean memorization; held-out fields scored against themselves give $1.00$, $0.96$ and $1.01$ at $N_{\text{train}}=1$, $10$ and $100$. Sampling time is per field on one A100.}
  \label{tab:baselines}
  \resizebox{\textwidth}{!}{%
  \begin{tabular}{lccccl}
    \toprule
    & \multicolumn{3}{c}{NN ratio at $N_{\text{train}}$} & & \\
    \cmidrule(lr){2-4}
    Method & $1$ & $10$ & $100$ & ms/sample & Outcome at $N_{\text{train}}=1$ \\
    \midrule
    Macrocanonical generator (ours)  & $0.89$  & $0.98$ & $0.99$ & $0.9$    & new fields \\
    Microcanonical synthesis         & $0.87$  & --     & --     & $9200$   & statistics pinned (no fluctuations) \\
    DDPM                             & $0.32$  & $0.58$ & $0.86$ & $107$    & memorizes training field \\
    WGAN-GP                          & $0.010$ & $0.33$ & $0.84$ & $0.09$   & memorizes training field \\
    WGAN-GP + translation aug.       & $0.24$  & --     & --     & $0.10$   & memorizes shifts of it \\
    Flow matching                    & $0.054$ & $0.24$ & $0.66$ & $36$     & memorizes training field \\
    Flow matching + translation aug. & $0.052$ & --     & --     & $36$     & memorizes training field \\
    \bottomrule
  \end{tabular}%
  }
\end{table}

The baselines improve as more independent fields become available, which is the expected signature of fitting the empirical data distribution: diffusion models in particular generalize only above a dataset-size threshold and reproduce their training data below it \cite{favero2025compositional}.
By $N_{\text{train}}=100$, WGAN-GP matches the mean feature vector at least as well as our generator (WPH distance $0.058$ against our $0.086$; held-out $0.037$), whereas the DDPM still has a WPH distance of $0.14$ and a KS mean-$p$ of $0.007$ (held-out $0.50$), with a per-coefficient scatter of only $0.28$ of the held-out level (ours: $0.67$).
The advantage of the macrocanonical generator is therefore confined to the regime in which fields are scarce, which is the regime that motivates the method.

\paragraph{Single training field at the paper's resolution.}
The $N_{\text{train}}=1$ row of Table~\ref{tab:nsweep} is limited by the $64{\times}64$ field size rather than by the training budget: increasing the budget from $1500$ to $5000$ iterations leaves the WPH distance ($0.20$ in both cases) and the NN ratio ($0.89$ against $0.87$) unchanged.
Repeating the single-field protocol on the $128{\times}128$ Quijote benchmark (one training slice, $100$ generator samples scored against $100$ held-out slices) gives a per-coefficient scatter ratio, the mean over WPH coefficients of the generated to the held-out across-sample standard deviation, of $1.14$ (held-out slices scored against themselves: $1.05$), an NN ratio of $1.00$ and a pixel-marginal KS distance of $0.015$ ($0.014$).
This is the result referred to in Section~\ref{sec:experiments}: the widths of the per-coefficient distributions are reproduced from a single $128{\times}128$ field, although their positions are not all matched at this training-set size (WPH distance $0.052$ against $0.021$ for held-out slices, KS mean-$p$ $0.15$ against $0.56$), as on the MHD suite.
}

{\ \section{Learning-rate, initialization-scale and input-distribution sweeps}
\label{app:sweeps}

Two hyperparameters govern whether training remains in the near-identity regime analyzed in Sections~\ref{sec:dynamics} and~\ref{sec:entropy}: the learning rate $\eta$ and the standard deviation $\sigma$ of the initial residual weights.
We sweep each in turn on the $64{\times}64$ MHD suite of Appendix~\ref{app:mhd64} ($N_{\text{train}}=100$, $600$ iterations at batch size $32$, multiplicative parameterization, all other settings as in Appendix~\ref{app:experiment-details}) and record three diagnostics on a fixed probe batch every $10$ iterations.
The sample diversity is the per-pixel standard deviation across the generated batch divided by that of the input batch, so a value near $1$ means the generator has not contracted the input distribution.
The per-step output change, $\|F_{\theta^{(t+1)}}(x)-F_{\theta^{(t)}}(x)\|/\|F_{\theta^{(t)}}(x)\|$, is the quantity that the linearization of Theorem~\ref{thm:preconditioned} requires to be small.
The input-Jacobian deviation, $\|\mathcal{J}_\theta(x)v-v\|/\|v\|$ for a random direction $v$ computed by finite differences, measures how far the map is from the identity and tracks the volume-preservation argument of Section~\ref{sec:entropy}.

\paragraph{Learning rate.}
For $\eta \in \{10^{-5}, 10^{-4}, 10^{-3}, 10^{-2}\}$ the median per-step output change grows as $0.03$, $0.085$, $0.14$ and $0.46$ (Table~\ref{tab:lr-sweep}).
At the paper's $\eta=10^{-4}$ the input-Jacobian deviation grows gradually from $0$ to $0.63$ over training, and the sample diversity never falls below $0.93$ and ends at $0.98$, the input level.
At $\eta=10^{-2}$ the first updates take the map far from the identity (input-Jacobian deviation above $300$ within the first $20$ iterations, diversity falling to $0.70$); this is the learning rate at which Figure~\ref{fig:ablation-inductive-bias} shows collapse after $100$ steps.
On this small problem the run later recovers its diversity, but it ends with a larger constraint error than $\eta=10^{-3}$.
The two smallest learning rates have not reached the constraints after $600$ iterations, which is why the paper's runs use $1500$ to $5000$ iterations at $\eta=10^{-4}$.

\begin{table}[h]
  \centering
  \small
  \caption{Learning-rate sweep ($64{\times}64$ MHD, $600$ iterations, all other settings fixed). The per-step change is the median relative output change on a fixed probe batch; the linearization of Theorem~\ref{thm:preconditioned} requires it to be small. The input-Jacobian deviation and diversity are given as the extreme value over training and, in parentheses, the value at the end of training; the loss is the mean over the last $50$ iterations.}
  \label{tab:lr-sweep}
  \resizebox{\textwidth}{!}{%
  \begin{tabular}{lcccc}
    \toprule
    $\eta$ & Median per-step change & Input-Jacobian deviation, max (final) & Diversity, min (final) & Loss \\
    \midrule
    $10^{-5}$          & $0.03$  & $0.40$ ($0.39$) & $0.94$ ($1.02$) & $2.0$ \\
    $10^{-4}$ (paper)  & $0.085$ & $0.65$ ($0.63$) & $0.93$ ($0.98$) & $1.2$ \\
    $10^{-3}$          & $0.14$  & $0.70$ ($0.69$) & $0.83$ ($0.99$) & $0.76$ \\
    $10^{-2}$          & $0.46$  & $322$ ($0.90$)  & $0.70$ ($0.98$) & $0.83$ \\
    \bottomrule
  \end{tabular}%
  }
\end{table}

\paragraph{Initialization scale.}
The paper draws the initial residual weights from $\mathcal{N}(0,\,\sigma^2)$ with $\sigma = 10^{-4}$; $\sigma$ controls how far the generator starts from the identity.
We sweep $\sigma \in \{10^{-5}, 10^{-4}, 10^{-3}, 10^{-2}, 3\times10^{-2}, 10^{-1}\}$ at $\eta=10^{-4}$ (Table~\ref{tab:sigma-sweep}).
Measured on the training inputs, the initial departure from the identity $\|F(x)-x\|/\|x\|$ runs from below $10^{-11}$ to $4.5$, so the sweep spans the range from the identity to a map far from it.
Training reaches the constraints from all six starting points: the mean loss over the final $50$ iterations ranges from $0.36$ to $0.85$, against a batch-to-batch scatter of $0.23$--$0.38$ within each run, with the largest $\sigma$ the worst, and the sample diversity at the end of training lies between $0.90$ and $0.96$ of the input's.
What changes is the trained map itself: its input-Jacobian deviation is $0.6$--$0.9$ for $\sigma \le 10^{-3}$ but $2.1$, $4.9$ and $11.1$ at $\sigma = 10^{-2}$, $3\times10^{-2}$ and $10^{-1}$, so a larger initialization leaves the trained map further from a perturbation of the identity, and the diversity is lowest at the largest $\sigma$.

\begin{table}[h]
  \centering
  \small
  \caption{Initialization-scale sweep ($64{\times}64$ MHD, $\eta=10^{-4}$, $600$ iterations, all other settings fixed). The loss is the mean over the last $50$ iterations; the diversity and input-Jacobian deviation are measured at the end of training.}
  \label{tab:sigma-sweep}
  \begin{tabular}{lcccc}
    \toprule
    $\sigma$ & $\|F(x)-x\|/\|x\|$ at initialization & Input-Jacobian deviation & Diversity & Loss \\
    \midrule
    $10^{-5}$           & $<10^{-11}$      & $0.81$ & $0.96$ & $0.51$ \\
    $10^{-4}$ (paper)   & $5\times10^{-5}$ & $0.61$ & $0.94$ & $0.70$ \\
    $10^{-3}$           & $0.046$          & $0.86$ & $0.96$ & $0.44$ \\
    $10^{-2}$           & $0.44$           & $2.1$  & $0.93$ & $0.36$ \\
    $3\times10^{-2}$    & $1.3$            & $4.9$  & $0.91$ & $0.49$ \\
    $10^{-1}$           & $4.5$            & $11.1$ & $0.90$ & $0.85$ \\
    \bottomrule
  \end{tabular}
\end{table}

\paragraph{Relation to the ablation of Figure~\ref{fig:ablation-inductive-bias}.}
The initialization sweep complements the ablation rather than repeating it.
In the ablation the residual shortcut is removed altogether and the transport collapses; with the shortcut in place the input still reaches the output even from a poor initialization, which is why the initialization scale has a wide margin around the paper's value.
The learning rate is the setting that decides whether training stays in the regime described by Theorem~\ref{thm:preconditioned}.

\paragraph{Input distribution.}
The paper's $p_0$ is a stationary Gaussian field with the power spectrum of the training pool (Appendix~\ref{app:experiment-details}).
To test the sensitivity to this choice, we repeated the $N_{\text{train}}=100$ training on the same MHD suite with an input that carries no power-spectrum information, white noise of the same variance, which is also stationary, so Corollary~\ref{cor:stationarity} still applies; each input was run twice, for $1000$ and $1200$ iterations.
Training is stable with either input and the sample variability is unchanged: relative to held-out fields, the per-pixel diversity is $1.02$ with the matched input and $1.06$ with white noise (held-out halves against each other: $1.02$), the mean pairwise distance between samples is $0.99$ and $1.02$ ($1.02$), and the per-coefficient scatter ratio is $0.66$ and $0.67$, scale by scale.
The results degrade measurably, however, in the highest-order statistic we measure: the relative error in the number of peaks above $1$, $2$ and $3\sigma$ of the held-out pixel distribution is $1.84$ and $3.56$ with white noise, against $0.43$ in both spectrum-matched runs.
The remaining held-out statistics are less sensitive on this small problem: the power-spectrum error is smaller with white noise ($0.12$ against $0.18$--$0.19$), as is the pixel-marginal KS distance ($0.08$--$0.12$ against $0.13$--$0.14$), and the mean-feature error differs by less than the run-to-run variation ($0.18$ against $0.13$ in one run, $0.13$ against $0.17$ in the other).
Matching the second-order structure of the target lets the generator spend its capacity on the higher-order structure rather than on building the correlations itself, which is where the white-noise input falls short, and we expect the gap to widen for steeper spectra, larger fields or shorter training.
We therefore recommend the spectrum-matched input used throughout the paper.}

\section{Wavelet phase harmonic summary statistics}
\vspace{-0.3cm}
\label{app:wph}

We use wavelet phase harmonic (WPH) coefficients \cite{mallat2020phase,zhang2021maximum} as the feature map $\phi$ in every experiment, computed by the \texttt{pywph} library with periodic boundary conditions (\texttt{pbc = True}).

WPH coefficients are constructed from the wavelet transform $\{W_{j,\ell} f\}$ of the field $f$ at $J$ dyadic scales and $L$ orientations, augmented with \emph{phase harmonics} $W_{j,\ell} f \mapsto |W_{j,\ell} f|\, e^{i p \arg W_{j,\ell} f}$ for an integer-valued exponent set $p$, and form translation-invariant second-order cross-moments across nearby scales (separated by up to $\Delta j$ octaves), orientations ($\Delta \ell$ rotations), and spatial translations ($\Delta n$ pixel shifts); see \cite{mallat2020phase} for the full vocabulary. This descriptor captures phase coupling that pure power-spectrum or wavelet-scattering moments do not represent, including filamentary structure, ridges, and small-scale alignment.

\paragraph{Per-benchmark configuration.}
For both stationary benchmarks of Section~\ref{sec:experiments} we use the same $J = 5$-level Morlet wavelet bank with $L = 4$ orientations and the \texttt{pywph} default scale and orientation separations $\Delta j = J - 1 = 4$ and $\Delta \ell = 2$, computed at the native field resolution $M = 128$.
The two benchmarks differ only in the spatial-shift order $\Delta n$ used in the cross-moments and in the set of phase-harmonic exponents (Table~\ref{tab:wph-config}).

For the \textbf{cosmological cosmic web (Quijote)} target we set $\Delta n = 0$ (cross-moments at the same position only) and pass an empty phase-harmonic exponent list (\texttt{p\_list} $= \emptyset$) to \texttt{pywph}'s \texttt{load\_model}, retaining the wavelet-scattering-like subset of the WPH family.
This choice is sufficient to capture the second-order non-Gaussian phase structure of the Quijote slices while keeping the coefficient vector compact.

For the \textbf{MHD turbulent fluid simulation} target we set $\Delta n = 1$ (adding a nearest-neighbour spatial shift to each cross-moment) and use \texttt{pywph}'s default phase-harmonic exponent list, yielding a richer WPH descriptor than the cosmic-web setting; we found this expanded coefficient vector to be necessary in order to match $\phi^\ast$ in the single-snapshot regime.

In both cases we use the library's automatic per-coefficient normalization (\texttt{norm = `auto'}), which rescales each coefficient by its empirical magnitude on the first call so that the contribution of every scale/orientation/exponent triple to the macrocanonical loss is comparable.
The first invocation of the operator on the training data primes this normalization cache; all subsequent calls during training reuse it.

\begin{table}[h]
  \centering
  \caption{WPH summary-statistic configuration per benchmark; symbols match \cite{mallat2020phase}.}
  \label{tab:wph-config}
  \begin{tabular}{lcccccc}
    \toprule
    Benchmark & $J$ & $L$ & $\Delta j$ & $\Delta \ell$ & $\Delta n$ & $p$-list \\
    \midrule
    Cosmological cosmic web (Quijote) & 5 & 4 & 4 & 2 & 0 & $\emptyset$ \\
    MHD turbulent fluid simulation    & 5 & 4 & 4 & 2 & 1 & \texttt{pywph} default \\
    \bottomrule
  \end{tabular}
\end{table}

\section{Per-coefficient WPH distribution test}
\vspace{-0.3cm}
\label{app:feature-dist}

\begin{figure}[h]
  \centering
  \demaybefig{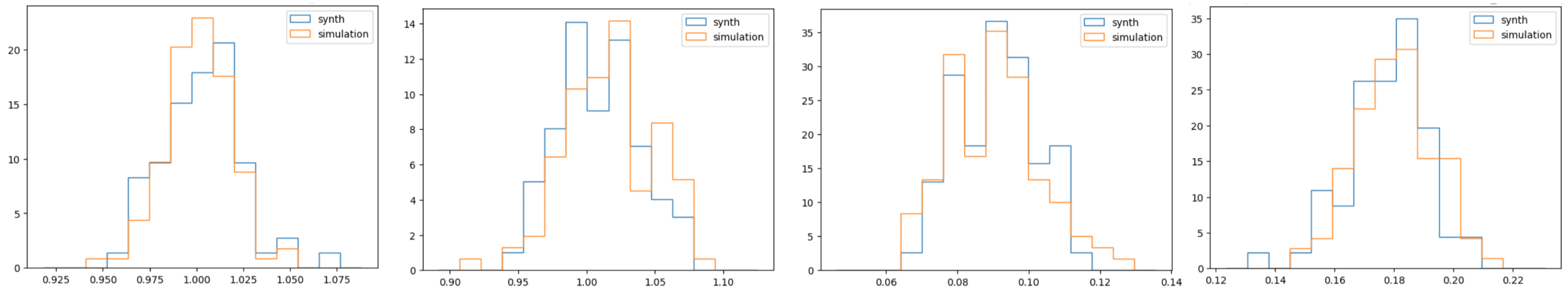}
  \caption{Across-sample distributions of four randomly chosen WPH coefficients from $100$ macrocanonical generator outputs (\emph{blue}) and $100$ held-out Quijote simulator slices (\emph{orange}). Microcanonical synthesis~\cite{bruna2018multiscale,zhang2021maximum} would instead collapse each panel to a Dirac mass at $\phi^\ast$.}
  \label{fig:wph-fluctuations}
\end{figure}

We back up the claim of Section~\ref{sec:cosmo} with a visual and a quantitative comparison.
After training the macrocanonical generator on $100$ Quijote cosmic-web slices against the empirical mean WPH vector $\phi^\ast$ alone---so the generator never sees per-realization fluctuations during training---we draw $100$ generator samples and compare the across-sample distribution of every WPH coefficient to that of $100$ held-out Quijote slices the network has never seen.
Figure~\ref{fig:wph-fluctuations} shows the comparison for four randomly chosen coefficients: the generator's per-realization distribution closely matches that of the held-out simulator pool in both mean and width, even though only the mean was supplied as a training target.
A complementary quantitative test runs a two-sample Kolmogorov--Smirnov (KS) test independently for every WPH coefficient; averaged over the full coefficient vector this yields a mean $p$-value of $\langle p \rangle = 0.52$, and the empirical distribution of $p$-values across coefficients is consistent with the uniform distribution on $[0, 1]$ expected under the null that both samples are drawn from the same underlying distribution.
This empirically confirms the entropy-preservation analysis of Section~\ref{sec:entropy}: scatter that is never directly supervised is nonetheless inherited from the input distribution and preserved by the residual identity-initialized generator.
Microcanonical synthesis~\cite{bruna2018multiscale,zhang2021maximum}, by construction, produces $\phi(x_i) = \phi^\ast$ for every sample and would therefore collapse each of the panels in Figure~\ref{fig:wph-fluctuations} to a Dirac mass at $\phi^\ast$, eliminating the per-feature variance that downstream simulation-based inference (Section~\ref{sec:sbi}) needs in order to learn a non-degenerate posterior.

\end{document}